\documentclass[sensors,article,submit,pdftex,moreauthors]{Definitions/mdpi} 

\firstpage{1} 
\pubvolume{1}
\issuenum{1}
\articlenumber{0}
\pubyear{2023}
\copyrightyear{2023}
\datereceived{ } 
\daterevised{ } 
\dateaccepted{ } 
\datepublished{ } 
\hreflink{https://doi.org/} 

\usepackage{pgfplots}
\usepackage{pgfplotstable}
\usepgfplotslibrary{colorbrewer}
\usepackage{booktabs}
\usepackage{subcaption} 
\usepackage{cellspace}
\usepackage{makecell}
\usepackage{mathpazo}

\pgfplotsset{every axis/.append style={label style={font=\fontsize{9}{12}\selectfont}}}

\definecolor{color1}{RGB}{168,221,181}
\definecolor{color2}{RGB}{123,204,196}
\definecolor{color3}{RGB}{78,179,211}
\definecolor{color4}{RGB}{43,140,190}
\definecolor{color5}{RGB}{8,88,158}

\definecolor{colorGraph1}{HTML}{d7191c}
\definecolor{colorGraph2}{HTML}{fdae61}
\definecolor{colorGraph3}{HTML}{ffffbf}
\definecolor{colorGraph4}{HTML}{abd9e9}
\definecolor{colorGraph5}{HTML}{2c7bb6}

\def\oM{\textbf{o}}
\def\oMwrong{\tilde{\oM}}
\def\iM{\textbf{i}}
\def\fm{f_m}
\def\dMm{d}
\def\fM{f_M}
\def\Hcam{H_{\text{cam}}}
\def\Hscene{H_{\text{scene}}}
\def\HcamML{H_{\text{cam}}^{\text{ML}}}
\def\HsceneML{H_{\text{scene}}^{\text{ML}}}
\def\X{X}
\def\dML{d_{\text{ML}}}
\def\hMLI{d_{\text{MLI}}}
\def\deltaUV{\Delta_{UV}}
\def\deltaXY{\Delta_{ST}}
\def\px{s_{px}}
\def\F{F}
\def\Fnew{F'}
\def\shift{S}
\def\shiftwrong{\tilde{\shift}}
\def\error{ERR}

\def\delt{\mathbf{\Delta}}
\def\deltwrong{\tilde{\delt}}
\def\errorshift{\error_{\shift}}
\def\errorowrong{\error_{\oMwrong}}
\def\errorswrong{\error_{\shiftwrong}}

\Title{Mind the Exit Pupil Gap: \\Revisiting the Intrinsics of a Standard Plenoptic Camera}

\TitleCitation{Title}

\Author{Tim Michels $^{1,*}$\orcidA{}, Daniel Mäckelmann $^{1}$ and Reinhard Koch $^{1}$}

\AuthorNames{Tim Michels, Daniel Mäckelmann and Reinhard Koch}

\AuthorCitation{Michels, T.; Mäckelmann, D.; Koch, R.}

\address{%
$^{1}$ \quad 
Department of Computer Science, Kiel University, 24118 Kiel, Germany

\corres{Correspondence: tim.michels@posteo.de}}

\abstract{
Among the common applications of plenoptic cameras are depth reconstruction and post-shot refocusing. 
These require a calibration relating the camera-side light field to that of the scene. Numerous methods with this goal have been developed based on thin lens models for the plenoptic camera's main lens and microlenses.
Our work addresses the often-overlooked role of the main lens exit pupil in these models and specifically in the decoding process of standard plenoptic camera (SPC) images.
We formally deduce the connection between the refocusing distance and the resampling parameter for the decoded light field and provide an analysis of the errors that arise when the exit pupil is not considered.
In addition, previous work is revisited with respect to the exit pupil's role and all theoretical results are validated through a ray-tracing-based simulation.
With the public release of the evaluated SPC designs alongside our simulation and experimental data we aim to contribute to a more accurate and nuanced understanding of plenoptic camera optics.}


\begin{document}
	
\crefname{equation}{equation}{equations}
\Crefname{equation}{Equation}{Equations}
\crefname{figure}{figure}{figures}
\Crefname{figure}{Figure}{Figures}

\setlength{\jot}{10pt}


\section{Introduction}\label{sec:Introduction}
\begin{figure}[h!]
	\centering
	\includegraphics[page=1]{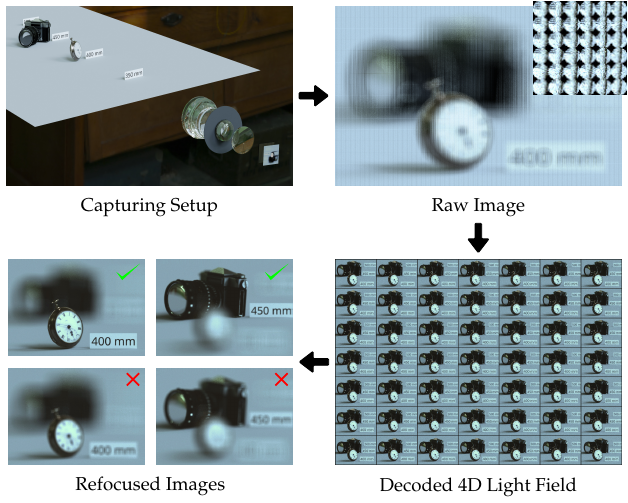}
	\caption{Exemplary pipeline for SPC post-shot refocusing: A scene is captured by a virtual SPC shown without housing. The resulting raw image consists of a large number of microlens images and is subsequently decoded into a 4D light field representation which is visualized by a subset of the sub-aperture images \cite{dansereau2013decoding}. By resampling the light field, a refocused image can be created \cite{ng2005lightfieldcamera}. The correctly focused images were created based on parameters considering the exit pupil as described in \cref{sec:SPC_Optics} while the slightly defocused image are results from the directly calculated parameters without exit pupil consideration based on \cite{pertuz2018focus}.}
	\label{fig:intro}
\end{figure}

Plenoptic cameras as initially described by Lippmann \cite{lippmann1908integralphoto} and Ives \cite{ives1928camera} combine a traditional camera with an additional microlens array (MLA) located between the main lens and the sensor. 
Over the years, two primary designs have been extensively studied and brought to market, the standard plenoptic camera (SPC) \cite{adelson1992single} \cite{ng2005lightfieldcamera} and the focused plenoptic camera (FPC) \cite{georgiev2006lightfieldcamdesign}\cite{perwass2012single}, which mainly differ in the microlens focus distance. Due to the earlier commercialization, a larger angular resolution and a simpler decoding process, the SPC still remains popular, even though it has certain disadvantages in terms of the spatial resolution and depth of field when compared to the multi-focus variant of the FPC \cite{perwass2012single}.
Classical applications for SPCs include depth reconstruction \cite{adelson1992single} and post-capture refocusing from single shots \cite{ng2005lightfieldcamera} and as a first step to achieve these, the raw 2D image of a plenoptic camera is usually de-multiplexed and resampled into a 4D light field \cite{dansereau2013decoding} as shown in \cref{fig:intro}. For this reparametrization 
procedure the knowledge about the exact position of each microlens image center (MIC) is crucial, as any inaccuracies in their locations can result in computational errors affecting the quality of the refocused images \cite{hahne2018baseline}.
Furthermore, 
a formal connection between the MICs and the plenoptic camera optical setup is required in order to relate the light field within the camera to the optical reality outside the camera, e.g. for finding the correct refocusing parameters for the desired object distance \cite{dansereau2013decoding}\cite{hahne2018baseline}.

Over the past two decades a number of studies have delved into the topic of processing plenoptic camera images, but often considered the MICs to be determined by the main lens center or its principal planes - a consequence of reducing the main lens to a simple thin lens. However, this assumption oversimplifies the actual optics involved. A more accurate representation acknowledges the role of the exit pupil in determining these image centers, as observed in studies by Hahne et al. \cite{hahne2018baseline}\cite{hahne2016refocusing}. Despite these advancements, the exit pupil is still often ignored in studies relating the light field within the camera to the 3D scene in front of the camera.

In this context, our work aims to highlight the importance of the exit pupil again. To this end first a paraxial model of the SPC under consideration of the exit pupil is described, directly relating the refocusing shift \cite{ng2005lightfieldcamera} to the object distance. The expected errors of the models ignoring the exit pupil is formally analyzed and later verified through a ray-tracing-based simulation of various plenoptic cameras in Blender \cite{blender} using real lens data. Subsequently, multiple works in the domain of plenoptic camera calibration are revisited and examined with respect to the need for a more complex lens model. More specifically, first the popular work of Dansereau et al. \cite{dansereau2013decoding} is revisited as well as the works of Zhang et al. \cite{zhang2018generic} and Monteiro et al. \cite{monteiro2019standard} building upon Dansereau's ideas. In these cases, one can conclude that the parameters of the respective calibration models are sufficiently general to permit the simplicity of a main lens model without considering the exit pupil. However, this does only hold true since these works do not require a specific interpretation of the model parameters.
For the work of Pertuz et al. \cite{pertuz2018focus} on the other hand, which also employs the decoding from \cite{dansereau2013decoding} for metric distance measurement, it is shown, that the oversimplified main lens model leads to an incorrect interpretation of their metric refocusing model parameters. In summary, our contributions are:
\begin{itemize}
	\item A formal deduction of the connection between object distance and sub-aperture image shift considering the exit pupil.
	\item A model for the errors resulting from ignoring the exit pupil in this relation.
	\item An analysis of the exit pupil's role in popular works on SPC calibration \cite{dansereau2013decoding}, \cite{zhang2018generic}, \cite{monteiro2019standard}, \cite{pertuz2018focus} and \cite{van2019focal}.
	\item Publicly available\footnote{\url{https://gitlab.com/ungetym/SPC-revisited} and \url{https://gitlab.com/ungetym/blender-camera-generator}} SPC 
	designs and a camera simulation framework based on Blender \cite{blender} supporting a large data base of lens designs and enabling a quick generation of new plenoptic camera setups.
\end{itemize}

\subsection{Related Work}\label{sec:RelatedWork}

\textbf{Plenoptic cameras:}
There are two primary design concepts of plenoptic cameras which have been extensively studied and brought to market. 
The first, known as the standard plenoptic camera (SPC) was described by Adelson and Wang \cite{adelson1992single} and later commercialized by Ng \cite{ng2005lightfieldcamera}. 
It requires the microlenses to be focused at infinity, i.e. the MLA-to-sensor separation must match the microlens focal length. 
Consequently, for a scene object placed at the SPC's focus distance, all sensor pixels behind a single microlens effectively capture a nearly identical segment of this object, albeit from slightly varied perspectives. This results in a large angular resolution, but a low spatial resolution.

\noindent Later, the focused plenoptic camera (FPC) was presented by Lumsdaine and Georgiev \cite{lumsdaine2009focused} and extended by Perwass and Wietzke \cite{perwass2012single} to feature multifocal MLAs for an extended depth of field. 
For the FPC, instead of directly dissecting the scene into its directional components, the microlenses are focused at the scene's virtual image inside the camera. 
This arrangement is advantageous in maintaining a greater portion of the conventional camera's spatial resolution albeit at the expense of a decrease in angular resolution compared to the SPC.

\noindent Despite the advantages of FPCs especially when combined with multifocal MLAs, we chose to focus this work on the SPC due to the straightforward image processing pipeline. More specific, to show the effect of the exit pupil, the application of post-shot refocusing is used throughout this work. And while this is a possible application for an FPC, the process involves a direct \cite{wanner2012globally} or indirect \cite{georgiev2010focused} depth estimation which can introduce artifacts affecting subsequent processing steps. In contrast, for a well configured SPC the post-shot refocusing only comprises of a demultiplexing step \cite{dansereau2013decoding} for the raw image with a subsequent shift-and-sum procedure for the resulting sub-aperture images. 
The simplicity of this pipeline reduces the quantity of artifacts resulting from complex interpolation and optimization steps and therefore allows a more direct analysis of the exit pupil's effects.
Note, however, that this work is intended to be the starting point for further research which will also analyze and improve FPC calibration algorithms suffering from the same model flaws as their SPC pendants. 

\noindent\textbf{SPC calibration:}
The calibration of plenoptic cameras plays a crucial role in relating the captured light field within the camera to the 3D world in front of the camera. 
To this end, Dansereau et al. \cite{dansereau2013decoding} present a method for demultiplexing the raw image of an SPC into a 2D array of sub-aperture images and using these for the geometric calibration. Due to its popularity and accessibility in form of a Matlab toolbox, this work is still being used as base for publications concerning the processing of SPC data. With respect to calibration models Zhang et al. \cite{zhang2018generic} and Monteiro et al. \cite{monteiro2019standard} modify the ideas of \cite{dansereau2013decoding} in order to associate a plenoptic camera with an equivalent multi-camera array. 
Both works make direct use of the decoding process proposed by Dansereau et al. \cite{dansereau2013decoding}. 
Pertuz et al. \cite{pertuz2018focus} also follow this approach and propose a focus-based metric depth estimation.

\noindent Due to the popularity of the demultiplexing process of \cite{dansereau2013decoding}, the revision of previous literature in \cref{sec:Revisiting} focuses on this method and the approaches based on it. Nevertheless, there is further work, which in part also reduces the main lens to a thin lens and thereby ignores the effects of the exit pupil. 
One such calibration approach, directly using line-features in the microlens images, is presented by Bok et al. \cite{bok2016geometric} and with a similar model Zhao et al. \cite{zhao2020metric} perform an SPC calibration based on plenoptic disc features. 
Thomasen et al. \cite{thomason2014calibration} as well as Suliga and Wrona \cite{suliga2018microlens} directly estimate the MLA pose and microlens pitch, but do not relate the captured light field to the scene-side light field.
Like the approaches related to \cite{dansereau2013decoding}, these works also assume the microlens images centers to be projections of the main lens center, i.e. the camera-side principal plane's center.
\noindent On the other hand Hahne et al. \cite{hahne2018baseline}\cite{hahne2016refocusing} describe the refocusing distance based on known main lens and MLA parameters under consideration of the exit pupil.
And further improvements to aspects of the calibration pipeline, which also acknowledge the exit pupil, are presented by Schambach et al. \cite{schambach2020microlens} increasing the MIC detection accuracy and Mignard-Debise and Ihrke \cite{mignard2019vignetting} analyzing the effect of vignetting on calibration models.

\noindent Of these works especially Hahne et al. \cite{hahne2016refocusing} consider the exit pupil and its connection to the microlens image geometry in a similar fashion to this work, but this is neither put into direct context of pre-existing calibration methods by a comparative evaluation nor does it provide an analysis of the expected errors resulting from oversimplified lens models. 
Nevertheless, to further validate our model, which establishes the refocusing distance in terms of the two-plane parametrization, its equivalence to the chief ray intersection model in \cite{hahne2016refocusing} is formally proven in \cref{subsec:equivalence}.

\noindent\textbf{FPC calibration:}
Despite this work focusing on SPCs as explained above, there is also relevant work in the related field of FPC calibration. Johannsen et al. \cite{johannsen2013calibration} describe a metric reprojection model for FPCs incorporating a radial distortion model. This is enhanced by Heinze et al. \cite{heinze2016automated} to also include the tilt and shift of the main lens as well as multi-focus MLAs. Further improvements to the distortion model are presented by Zeller et al. \cite{zeller2016metric}.
All these approaches are based on the reconstruction of the virtual scene between MLA and main lens and associate these virtual 3d points to the known scene points.
In contrast, Noury et al. \cite{noury2017light} propose an approach directly working on the microlens images, i.e. associating the scene points with their projections on the sensor without the intermediate step of calculating virtual depths. This method, however, is limited to single focus FPCs and models the microlenses as simple pinholes.
Nousias et al. \cite{nousias2017corner} on the other hand feature a more complete microlens model and directly include the estimation of multiple microlens focal lengths into their approach.
And Wang et al. \cite{wang2018virtual} present a two-step model including a forward projection from the scene into the camera and a second projection from the virtual image to the sensor.
More recently, Labussiere et al. \cite{labussiere2022leveraging} proposed a simultaneous calibration of the different microlens types in a multi-focus plenoptic camera by incorporating defocus blur into the features used for the parameter optimization.

\noindent None of the listed methods for FPC calibration directly consider the exit pupil and while most of these works, which require the identification of MICs, incorporate a scaling between the grid of microlens centers and the grid of MICs, this is usually a result of projecting the main lens center, i.e. the center of the camera-side principal plane, through the microlens centers. 
However, as observed by Hahne et al. \cite{hahne2018baseline}\cite{hahne2016refocusing} for SPCs and confirmed in \cref{sec:Evaluation}, the MICs result from a projection of the exit pupil's center instead. Thus using the distance between the simplified main lens plane and the MLA for the image formation model as well as the calculation of MICs could inadvertently reduce the degrees of freedom of the model. And while this might be desired in terms of an increased stability during the parameter optimization, this reduction should also be analyzed for FPC models. However, for reasons of clarity and comprehensibility we decided against also including the topic of FPC calibration into this work and leave this for future work.

\noindent\textbf{Lens models and simulation:}
In the domain of ray tracing-based camera simulation, realistic main lens models which consider all lens components and their respective properties have been used for over two decades, either explicitly 
by direct modeling as in Kolb et al. \cite{kolb1995lensmodelforsimulation} and Wu et al. \cite{wu2011accurate} or implicitly via learned black-box lens models as proposed in Zheng et al. \cite{zheng2017neurolens}. 
Regarding plenoptic cameras, most previous work uses oversimplified models for rendering, such as pinhole cameras or multi camera arrays modeling the MLA, but without a model for the main lens \cite{fleischmann2014plenoptic}\cite{zhang2015reconstruction}\cite{liang2015simuWOmainlens}.
More recently, Nürnberg et al. \cite{nurnberg2019simulation} as well as our group \cite{michels2018simulation} provided simulations of plenoptic cameras without oversimplifying the main lens. Due to familiarity we extended our previous work for the synthetic experiments.

\subsection{Organization}\label{sec:Organization}

In \cref{sec:SPC_Optics} first the general lens model and two-plane parametrization are explained before deducing the refocusing model under consideration of the exit pupil. Subsequently, \cref{sec:erroranalysis} provides a formal analysis of the expected errors when dismissing the exit pupil. In \cref{sec:Revisiting} previous works are revisited with focus on the need for more complex lens models. Finally, our deductions are validated with synthetic experiments in \cref{sec:Evaluation}.

\section{SPC Optics}\label{sec:SPC_Optics}

\def\Vec#1#2#3{\begin{bmatrix}#1 \\#2 \\#3\end{bmatrix}}

\subsection{Preliminaries - Lens Models}\label{subsec:Preliminaries}
\begin{figure}[h]
	\centering
	\includegraphics[page=1]{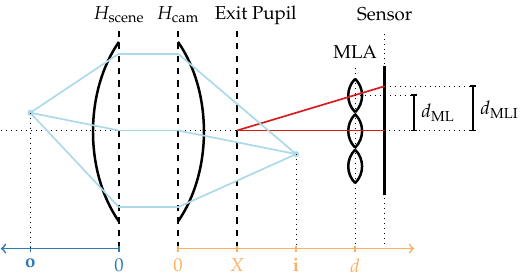}
	\caption{Plenoptic camera modeled by a thick main lens combined with a thin lens MLA. The microlens pitch is described by $\dML$ and the distance between neighboring microlens image centers (MICs) is denoted as $\hMLI$. Furthermore, $\X$ describes the distance between the exit pupil and the camera-side principal plane and $\dMm$ is the distance between $\Hcam$ and the MLA. A complete notation overview is given in \cref{append:notation}.}
	\label{fig:lensmodel}
\end{figure}

The thin lens model describes a lens by assuming it to be infinitely thin and only refracting light at a single lens plane. The relation between the real scene and the lens image in this model is described by the equation
\begin{equation}\label{eq:thinlens}
	\frac{1}{\fM} = \frac{1}{\oM} + \frac{1}{\iM}
\end{equation}
where $\fM$ is the focal length of the lens, $\oM$ is the object distance, and $\iM$ is the image distance, both measured from the refraction plane.
This concept can be extended to a thick lens model by expanding the refraction plane into two principal planes, $\Hscene$ and $\Hcam$, between which a traced light ray is considered to run parallel to the optical axis \cite{born2013principles}\cite{smith2008modern}.
Furthermore, a combination of thick lenses, such as the main lens of a plenoptic camera, can again be represented as a single thick lens \cite{smith2008modern}.
And as visualized in \cref{fig:lensmodel}, the object and image distances, $\oM$ and $\iM$, are then measured based on the positions of the principal planes and \cref{eq:thinlens} remains valid.
\\
In addition to this model one can consider the exit pupil, i.e. the image of the aperture stop viewed towards the image plane. It defines the size and location of the virtual aperture in the optical system \cite{smith2008modern} and, as pointed out by Hahne et al. \cite{hahne2018baseline}\cite{hahne2016refocusing}, determines the positions of the microlens image centers (MIC) on the sensor. As empirically shown in \cref{sec:erroranalysis} the exit pupil and $\Hcam$ rarely coincide and accordingly a systematic error could be introduced when a plenoptic camera image is de-multiplexed based on MICs incorrectly estimated under the premise that the main lens follows the thin lens model without considering the exit pupil.

\subsection{Preliminaries - Light Field Parametrization}\label{subsec:LFparam}

\begin{figure}[h!]
	\centering
	\includegraphics[page=1]{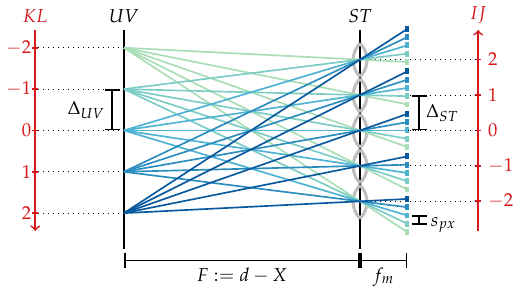}
	\caption{Integer (red) and metric (black) two plane parametrization of the light field. Here, $\px$ describes the size of a sensor pixel and $\fm$ the focal length of a microlens which for an SPC coincides with the distance between MLA and sensor.}
	\label{fig:LFparam}
\end{figure}

Despite being a standard tool when working with light field data, the two-plane parametrization as described by Levoy and Hanrahan \cite{levoy2023light} and used in various popular works including the work of Dansereau et al. \cite{dansereau2013decoding} and Ng et al. \cite{ng2005lightfieldcamera} are reiterated in this section for two reasons. First, the previous descriptions do not consider the exit pupil, and second, the literature is not consistent in terms of the underlying data representation. While \cite{dansereau2013decoding} uses the raw camera image, indexed by integer pixel coordinates, to base the description on, \cite{ng2005lightfieldcamera} assumes known metric coordinates for every pixel. We follow the approach of \cite{dansereau2013decoding} to facilitate the reproduction of our results.

Given an SPC following the thick lens model with an exit pupil as visualized in \cref{fig:lensmodel}, the light field inside this camera can be parameterized using two planes 
- the MLA, which serves as virtual sensor plane, and the exit pupil plane, which can be interpreted as virtual lens plane. By following the decoding process of Dansereau et al. \cite{dansereau2013decoding} the 4D light field can be parametrized as	$L_{\F}(i, j, k, l)$ with integer indices $(k,l)$ for the uniformly sampled sub-aperture image and $(i,j)$ for the pixel coordinates in that image. 

The corresponding metric parametrization $\tilde{L}_F(s,t,u,v)$ describes the intensity of light captured at the MLA plane point $(s,t,\dMm)$ coming from the exit plane point $(u,v,\X)$. In accordance with Ng et al. \cite{ng2005lightfieldcamera}\cite{ng2005fourier} \footnote{
	Note, that Ng et al. implicitly assume $\X=0$ since the main lens in that work is modeled by a thin lens. While this is approximately correct for the two tested main lenses in \cite{ng2005lightfieldcamera}, a Zeiss Planar T* 2/80 with $\X=7.8mm=0.098\cdot \fM$ and a Zeiss Sonnar T* 2.8/140 with $\X=-4mm=-0.029\cdot \fM$ (compare \cite{zeissPlanar}\cite{zeissSonnar}), \cref{sec:erroranalysis} shows, that this assumption does not hold in general.} 
the distance between these two parametrization planes is denoted as $F:=\dMm-\X$. 
To calculate the metric parametrization from a given integer parametrization, note, that the pixel pitch $\deltaXY$ of the virtual sensor, i.e. the step size in the $ST$-Plane, corresponds to the microlens pitch, i.e. $\deltaXY=\dML$ and, as shown in \cref{fig:LFparam}, the step size in the virtual lens plane, i.e. the $UV$-plane, can be calculated by means of the triangle equality as $\deltaUV = \frac{\px \cdot\F}{\fm}$ where $\px$ and $\fm$ denote the pixel size and the microlens focal length. 
With these step sizes the light field parametrized in metric coordinates $(s,t,u,v)$ is given by
\begin{equation}
	\tilde{L}_F(s,t,u,v)=L_{\F}\left(\frac{s}{\deltaXY},\frac{t}{\deltaXY},\frac{u}{\deltaUV},\frac{v}{\deltaUV}\right).
\end{equation}
Note, that the metric coordinates $(s,t,u,v)$ might not be integer multiples of their respective step sizes and accordingly, querying the corresponding values from the integer parametrization $L_{\F}$ could require additional interpolation steps.

\noindent With the described light field parametrization one can now reproduce the resampling steps necessary to refocus the image by moving the virtual sensor plane under consideration of the exit pupil position.

\subsection{Light Field Refocusing with Exit Pupil}\label{subsec:refocus}

\begin{figure}[h!]
	\centering
	\includegraphics[page=1]{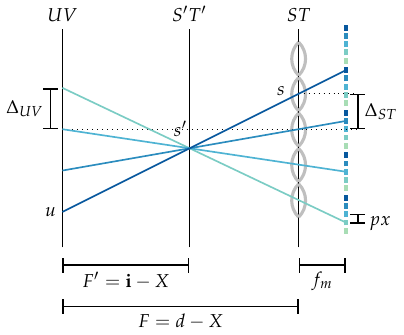}
	\caption{Light field refocusing via shift of the virtual sensor, i.e. the $ST$-plane is moved to the image distance $\iM$. A ray $(s',u)$ can be associated with a ray $(s,u)$ by means of the triangle equality, i.e. $s=u+\frac{F}{F'}(s'-u)$.}
	\label{fig:LFrefocus}
\end{figure}

In order to refocus the virtual sensor image onto an object at distance $\oM$, measured from $\Hscene$, this virtual sensor now needs to be placed at the distance $\iM$, measured from $\Hcam$, according to the thin lens \cref{eq:thinlens}.
This corresponds to a distance $\Fnew:=\iM-\X$ between the $UV$-plane (exit pupil) and the virtual sensor as visualized in \cref{fig:LFrefocus}.
By defining the refocusing parameter $\alpha=\frac{\Fnew}{\F}$ as in \cite{ng2005lightfieldcamera}, the thin lens equation can be applied to deduce
\begin{equation}\label{eq:alpha}
	\alpha=\frac{\Fnew}{\F}=\frac{\frac{\oM\cdot\fM}{\oM-\fM}-\X}{\dMm-\X}=\frac{\oM\cdot(\fM-\X)+\fM\cdot\X}{(\oM-\fM)(\dMm-\X)}.
\end{equation}

Given an integer 4D light field $L_{\F}(i,j,k,l)$ based on sub-aperture images as in \cite{dansereau2013decoding}, the relationship between the virtual sensor movement specified by $\alpha$ and the resulting disparity at the original $ST$-plane is described in the following. While the general deduction is similar to \cite{ng2005lightfieldcamera}, the following calculations are based on integer indexing for the purpose of reproducibility.

As shown in \cref{fig:LFrefocus}, the metric light field value $\tilde{L}_{\Fnew}(s',t',u,v)$ for the modified sensor plane placed at the distance $\Fnew$ from the exit pupil can be calculated as
\begin{align}
	&\tilde{L}_{\Fnew}(s',t',u,v)=\tilde{L}_{\F}\left(u+\frac{s'-u}{\alpha},v+\frac{t'-v}{\alpha},u,v\right) \nonumber\\
	&=\tilde{L}_{\F}\left(u\left(\frac{\alpha-1}{\alpha}\right)+\frac{s'}{\alpha},v\left(\frac{\alpha-1}{\alpha}\right)+\frac{t'}{\alpha},u,v\right)
\end{align}

Ignoring the image magnification introduced by the virtual sensor movement, i.e. setting the step size for the $S'T'$-Plane to $\alpha\cdot\deltaXY$ and defining $\delt:=\frac{\deltaUV}{\deltaXY}$, one can deduce, that the integer parametrization $L_{\Fnew}(i,j,k,l)$ for the modified sensor plane corresponds to
\begin{equation*}
	L_\F\left(k\cdot\delt\left(1-\frac{1}{\alpha}\right) + i, l\cdot\delt\left(1-\frac{1}{\alpha}\right) + j, k, l\right).
\end{equation*}
At this point, the pixel shift $\shift$ between neighboring sub-aperture images required to refocus onto the desired distance $\oM$ can be calculated given the value $\alpha$ as
\begin{equation}\label{eq:shift_alpha}
	\shift(\alpha)=\delt\cdot\left(1-\frac{1}{\alpha}\right)
\end{equation}
and plugging \cref{eq:alpha} into \cref{eq:shift_alpha} yields the direct relation between the object distance $\oM$ and the disparity $\shift$ as
\begin{align}\label{eq:S_correct}
	\shift(\oM)= \delt\cdot\frac{\oM(\fM-\dMm)+\fM\cdot\dMm }{\oM(\fM-\X) + \fM\cdot\X}.
\end{align}
This model can easily be reverted to calculate the object or refocusing distance based on a given sub-aperture image shift via
\begin{align}\label{eq:o_correct}
	\oM(\shift)= \frac{\fM\cdot(\dMm\cdot\delt-\shift\cdot\X)}{\shift\cdot(\fM-\X)-\delt(\fM-\dMm)}.
\end{align}

\section{Error Analysis}\label{sec:erroranalysis}
\begin{figure*}[h]
	\centering
	\includegraphics[page=1]{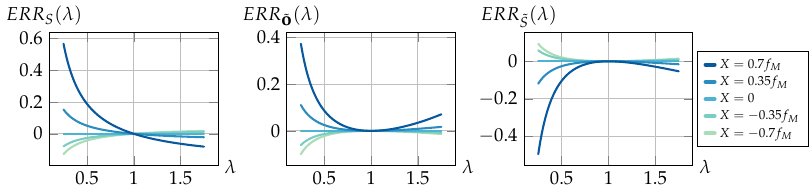}
	\caption{Left: Relative shift error based on $\lambda=\frac{\oM}{\oM_f}$. Mid/right: The two cases of relative object distance errors for the assumed camera focused at a finite distance which is met for a relative distance of $\frac{\oM}{\oM_f}=1$. Negative error values indicate an underestimation of the ground truth value while positive errors represent an overestimation. 
	}
	\label{fig:error_example}
\end{figure*}

In the following the error that can be expected by ignoring the exit pupil, i.e. setting $\X = 0$, is analyzed. First, we define the scaling between the $ST$ and $UV$ plane in under this assumption as
\begin{equation}
	\deltwrong := \mathbf{\Delta_{X=0}} = \frac{\px \cdot \dMm}{\fm\cdot\dML}
\end{equation}
and calculate the pixel disparity based on \cref{eq:S_correct} as
\begin{align}\label{eq:S_incorrect}
	\shiftwrong(\oM) :=\shift_{\X=0}(\oM)= \deltwrong\cdot\frac{\oM(\fM-\dMm)+\fM\cdot\dMm }{\oM\cdot \fM}
\end{align}
as well as the object distance which can be simplified to
\begin{align}\label{eq:o_incorrect}
	\oMwrong(\shift) := \oM_{X=0}(\shift)= \frac{\fM\cdot\dMm\cdot\deltwrong}{\shift\cdot\fM-\deltwrong(\fM-\dMm)}.
\end{align}
The relative error of the shift is then calculated by
\begin{align}\label{eq:error_S}
	\errorshift(\oM) := \frac{\shiftwrong(\oM)-\shift(\oM)}{\shift(\oM)} = \X\cdot\frac{\dMm\cdot\fM-\oM(\dMm-\fM)}{\oM\cdot\fM(\dMm-\X)}
\end{align}
and by describing $\oM$ as a multiple of the focus distance $\oM_f$, i.e. $\oM = \lambda\cdot\oM_f$, one obtains
\begin{align}\label{eq:error_S_lambda}
	\errorshift(\lambda):=\errorshift(\lambda\cdot\oM_f) = \frac{\X(\lambda-1)}{\lambda\oM_f\cdot\left(\frac{\X}{\dMm}-1\right)}.
\end{align}
For the error of the object or refocusing distance two cases are analyzed.
First, it is assumed, that the correct shift $\shift$ corresponding to the ground truth $\oM$ is given, but in a second step the oversimplified model of \cref{eq:o_incorrect} is used to calculate the object distance.
This error can be found in applications measuring the correct shift, e.g. by repeatedly refocusing an image and subsequently using the incorrect object distance calculation in order to estimate the associated metric distances in the scene.
This error can be formulated as
\begin{align}\label{eq:error_o}
	&\errorowrong(\oM) := \frac{\oMwrong(\shift(\oM))-\oM}{\oM} = -\frac{\X(\dMm-\fM)+\frac{\oM}{\oM_f}\X(\fM-\dMm)+\X\fM\dMm\left(\frac{1}{\oM_f}-\frac{1}{\oM}\right)}{\fM(\dMm-\X)+\frac{\oM}{\oM_f}\X(\fM-\dMm)+\X\fM\dMm\frac{1}{\oM_f}}
\end{align}
and using $\oM = \lambda\cdot\oM_f$ again, one obtains
\begin{equation}\label{eq:error_o_lambda}
	\errorowrong(\lambda):=\errorowrong(\lambda\oM_f) = \frac{\X(\lambda-1)^2}{\lambda\oM_f\left(1-\frac{\X}{\dMm}\right)-\X(\lambda-1)\lambda}.
\end{equation}
The second case assumes an incorrectly calculated shift $\shift$ based on \cref{eq:S_incorrect} which is subsequently used to refocus an image with the refocusing algorithm complying with the correct object distance estimation in \cref{eq:o_correct}.
This type of error is given by
\begin{align}\label{eq:error_o2}
	&\errorswrong(\oM) := \frac{\oM(\shiftwrong(\oM))-\oM}{\oM} = \frac{\fM\delt-\fM\X\deltwrong\left(\frac{1}{\oM}-\frac{1}{\oM_f}\right)}{\deltwrong\left(1-\frac{\oM}{\oM_f}\right)(\fM-X)+\delt\fM\frac{\oM}{\oM_f}}-1
\end{align}
and after substituting $\oM = \lambda\cdot\oM_f$ the error can be formulated by
\begin{align}\label{eq:error_o2_lambda}
	\errorswrong(\lambda):=\errorswrong(\lambda\oM_f) = \frac{X(\lambda-1)^2}{\lambda\oM_f\left(\frac{\X}{\fM}-1\right)-\lambda^2\X}.
\end{align}
Now assuming a camera with a focal length of $\fM = 100mm$ focused at a finite distance \cref{fig:error_example} shows exemplary error values for different values of $\X$ relative to the focal length.

\noindent The visualization shows that all errors diverge for $\lambda \to 0$ with a rate depending on the positional relationship between the exit pupil and the principal plane $\Hcam$.
Beyond the focus distance at $\lambda =1$ the object errors again diverge while the shift error converges according to $\errorshift(\lambda) \underset{\lambda\to\infty}{\longrightarrow} \frac{\X}{\oM_f(\X/\dMm-1)}$. Note, that these graphs present an ideal refocusing case free of aliasing artifacts and limiting optical properties such as the depth of field. Hence, later experiments only verify a section of these results within the respective physical and image processing limits.

\noindent In summary these examples show a large deviation between the estimated 
refocusing distances in models with and without consideration of the exit pupil whenever there is a non-zero distance $\X$ between the exit pupil and $\Hcam$. This leads to the question, how prevalent a significant $\X \neq 0$ is in off-the-shelf main lenses. To answer this question, the data of 866 DSLR lenses listed by Claff \cite{claff} was collected and $X$ as well as $\fM$ were calculated via paraxial ray-tracing for each lens.
\begin{figure}[h]
	\centering
	\includegraphics[page=1]{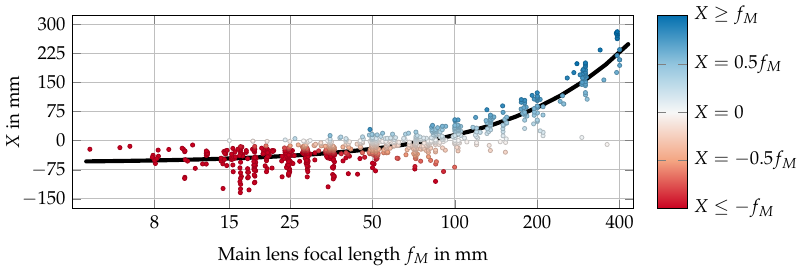}
	\caption{Distances $\X$ between exit pupil and principal plane $\Hcam$ for 866 lenses \cite{claff} sorted by focal length. The black line represents the linear model fitted to the data and the colors of the data points indicate the relationship $\X/\fM$. Note that the horizontal axis uses a logarithmic scaling due to the large number of lenses with a focal length below 100mm.}
	\label{fig:lenslist}
\end{figure}
The resulting data in \Cref{fig:lenslist} shows a nearly linear connection between the focal length of a lens and the distance $\X$ with a Pearson correlation coefficient of $0.8994$. Fitting a linear model to this data results in the non-zero function $\X(\fM)=0.7108\cdot\fM-56.5546$ with a coefficient of determination $R^2=0.8089$. Further examination shows, that only a small subset of 62 of the 866 lenses exhibits values for $\X$ below 5\% of the focal length, i.e. $|\X|<0.05\fM$. On the other hand, for 627 lenses the deviation is larger than $|\X|>0.25\fM$ and 444 lenses even have values $|\X|>0.5\fM$. Overall this data shows, that the assumption of $\X \approx 0$ is usually not met by reality. Therefore, the exit pupil should be considered when relating the camera-side light field to the scene's light field.

\section{Revisiting SPC Methods}\label{sec:Revisiting}
In this section several previous works are examined with respect to the exit pupil's role for the respective model deductions.

\subsection{Equivalent Ray Model}\label{subsec:equivalence}
First, the equivalence between the refocusing model in \cref{eq:o_correct} and the ray intersection model presented by Hahne et al. \cite{hahne2016refocusing} is proven. Instead of basing the model on decoding scheme of Dansereau et al. \cite{dansereau2013decoding} in \cite{hahne2016refocusing} an approach building upon the intersection of chief rays is presented in order to calculate the refocusing distance for a resampling of the raw plenoptic camera image. A comprehensive notation transfer into our setup is given in \cref{append:hahnenotation}.

\begin{figure}[h!]
	\centering
	\includegraphics[page=1]{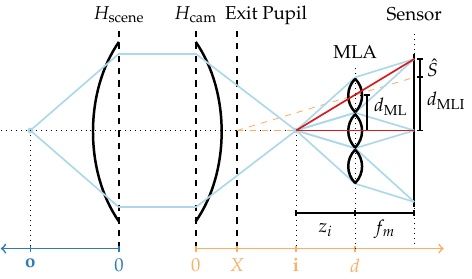}
	\caption{Image formation (light blue) for an object point located at distance $\oM$ from $\Hscene$. The image of this point, located at distance $\iM$ from $\Hcam$, is seen by multiple microlenses and its projections onto the sensor have a metric disparity of $\hat{\shift}$. In order to determine the object distance $\oM$, Hahne et al. \cite{hahne2016refocusing} propose to intersect ray functions (red) from two of the images and transfer the resulting image distance $\iM$ to the scene via the thin lens equation.}
	\label{fig:lensmodelhahne}
\end{figure}

The basic idea of \cite{hahne2016refocusing} as depicted in \cref{fig:lensmodelhahne} is the selection of two pixels on the sensor, which show scene points from the desired focus plane, and tracing rays from these through the respective microlens centers. The resulting camera-side intersection determines the distance of the virtual image inside the camera from the main lens and accordingly, the thin lens equation can be applied in order to calculate the corresponding object or refocusing distance. Without loss of generality, the following calculations will assume an MLA with one microlens center located on the main lens' optical axis.

Given a sub-aperture image shift $\shift$ in pixel as in \cref{subsec:refocus}, this translates to a metric pixel disparity $\hat{\shift}$ on the sensor by
\begin{equation}\label{eq:tildeshift}
	\hat{\shift} = -\frac{\px}{\shift}
\end{equation} 
with the sign flip resulting from the differing conventions used throughout this work and \cite{hahne2016refocusing}.
Under the premise of a well configured plenoptic camera with a regular microlens grid, any two pixels from neighboring microlenses with a disparity of $\hat{\shift}$ can be chosen to calculate the image distance.
To simplify the calculations, the first ray is chosen to run along the optical axis as shown in \cref{fig:lensmodelhahne} and the second ray is based on the pixel at sensor position $\hMLI + \hat{\shift}$ going through the neighboring microlens center.
According to \cite{hahne2016refocusing} (compare \cref{append:hahnenotation}) this leads to two ray functions
\begin{equation}
	f(z)=0\quad\text{and}\quad \tilde{f}(z)=\frac{\dML-(\hMLI + \hat{\shift})}{\fm}\cdot z+\dML
\end{equation}
which intersect at
\begin{equation}
	z_i=-\frac{\dML\cdot\fm}{\dML-(\hMLI+\hat{\shift})}.
\end{equation}
The crucial part, that sets \cite{hahne2016refocusing} apart from the works reviewed in the following sections, is the correct calculation of the microlens image center distance $\hMLI$ based on the exit pupil (compare the calculation of $u_{c,j}$ in \cref{table:hahne_transfer2}) via 
\begin{equation}
	\hMLI = \frac{\dML}{\dMm-\X}\cdot \fm + \dML
\end{equation}
which yields
\begin{align}
	z_i&=-\frac{\dML\fm}{\dML-(\frac{\dML}{\dMm-\X}\cdot \fm + \dML+\hat{\shift})}\nonumber\\
	&=\frac{(\dMm-\X)\dML\fm}{(\dMm-\X)\hat{\shift}+\fm\dML}.
\end{align}
This intersection results in the image distance $\iM=\dMm-z_i$ measured from $\Hcam$ and can be used to calculate the object distance via the thin lens equation by
\begin{align}
	\oM &= \left(\frac{1}{\fM}-\frac{1}{\iM}\right)^{-1}= \left(\frac{1}{\fM}-\frac{1}{\dMm-z_i}\right)^{-1}\nonumber\\
	&=\frac{\fM\cdot\dMm-\fM\cdot z_i}{(\dMm-\fM)-z_i}\nonumber\\ 
	&= \frac{\fM\cdot\dMm\cdot((\dMm-\X)\hat{\shift}+\fm\dML)-\fM\cdot (\dMm-\X)\dML\fm}{(\dMm-\fM)\cdot((\dMm-\X)\hat{\shift}+\fm\dML)-(\dMm-\X)\dML\fm}\nonumber\\ 
	&=\frac{\fM\cdot\dMm\cdot\frac{\dMm-\X}{\fm\dML}\hat{\shift}+\fM\X}{(\dMm-\fM)\cdot\frac{\dMm-\X}{\fm\dML}\hat{\shift}-\fM+\X}\nonumber\\ 
	&=\frac{\fM\cdot(\dMm\cdot\delt-\shift\cdot\X)}{\shift\cdot(\fM-\X)-\delt(\fM-\dMm)}
\end{align}
where \cref{eq:tildeshift} and the definition $\delt:=\frac{\deltaUV}{\deltaXY}=\frac{\px(\dMm-\X)}{\fm\dML}$ from \cref{sec:SPC_Optics} are used in the last step.
This equation equals the previously deduced \cref{eq:o_correct} and thus proves the equivalence of both models. 

\subsection{Light Field Decoding and SPC Calibration}\label{subsec:DansAndCo}
This section revisits the popular decoding and calibration theme presented by Dansereau et al. \cite{dansereau2013decoding}. In that work the raw plenoptic camera image is first de-multiplexed into an integer indexed two plane parametrization $L(i,j,k,l)$. These indices are then transformed into metric rays and propagated through the main lens. The combination of these steps yields an intrinsics matrix
\begin{equation}\label{eq:dans}
	\begin{bmatrix}
		s\\t\\u\\v\\1
	\end{bmatrix} = \begin{bmatrix}
		H_{1,1} & 0 & H{1,3} & 0 & H_{1,5}\\
		0 & H_{2,2} & 0 & H_{2,4} & H_{2,5} \\
		H_{3,1} & 0 & H{3,3} & 0 & H_{3,5}\\
		0 & H_{4,2} & 0 & H_{4,4} & H_{4,5} \\
		0 & 0 & 0 & 0 & 1
	\end{bmatrix} \cdot 
	\begin{bmatrix}
		i\\j\\k\\l\\1
	\end{bmatrix}
\end{equation}
associating the integer indices directly with metric coordinates $(s,t,u,v)$ for the scene-side light field. Note, that these do not correspond to the equally named coordinates in \cref{sec:SPC_Optics} which describe the camera-side light field coordinates before propagating them through the main lens.

\begin{figure}[h!]
	\centering
	\includegraphics[page=1]{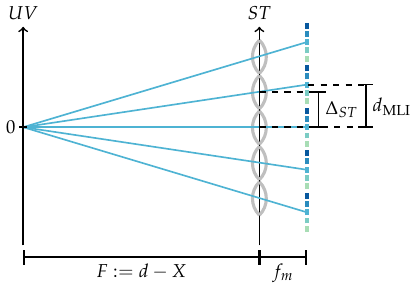}
	\caption{The central sub-aperture image consists of the MICs, i.e. the image of the aperture center viewed across all microlenses. These MICs originate from the center of the UV plane, i.e. the exit pupil. Accordingly, the distance between neighboring MICs is given by the triangle equality via $\hMLI=\deltaXY\cdot(\fm+\F)/\F=\deltaXY\cdot(1+\fm/(\dMm-\X))$.}
	\label{fig:MICs}
\end{figure}

The relevant step with respect to the gap between the exit pupil and the principal plane in this process is the division of the integer indices by the respective spatial frequencies of the pixels and microlenses via the matrix $\mathbf{H}_{abs}^\theta$. 
As explained in \cref{subsec:Preliminaries} the grid of MICs corresponds to the scaled grid of microlens centers. Accordingly, the sampling rate for the microlens plane has to be scaled down, or equivalently the pixel sampling rate has to be scaled up by the inverse factor.
Dansereau et al. \cite{dansereau2013decoding} acknowledge this fact and choose the second option, 
by introducing a scaling factor, which in our notation (compare \cref{append:dansnotation}) corresponds to \begin{equation}
	M_\text{proj}=\left(1+\frac{\fm}{\dMm}\right)^{-1}.
\end{equation} This scaling, however, assumes a projection center at the main lens principal plane. Using the exit pupil instead, the correct rescaling is given by
\begin{equation}
	M_\text{proj}=\left(1+\frac{\fm}{\dMm-\X}\right)^{-1}
\end{equation}
as visualized in \cref{fig:MICs}.

Fortunately, due to the overall formulation of the intrinsics as an end-to-end ray transformation, this slight change is not relevant for the calibration results, since neither $\fm$ nor $\dMm$ is directly estimated. Instead, the factor $M_\text{proj}$ contributes to the intrinsic variables $H_{1,1}$ to $H_{4,4}$ and repeating the deduction of $H$ with the correct scaling leads to the same general form for the intrinsics matrix as in \cref{eq:dans}.

Similar cases of general parameters compensating for the model inaccuracies are presented by Monteiro et al. \cite{monteiro2019standard} and Zhang et al. \cite{zhang2018generic}. Both make use of the same decoding process as \cite{dansereau2013decoding} and build upon the idea of directly relating the camera-side and scene light field. Monteiro et al. \cite{monteiro2019standard} slightly reduce the intrinsics matrix shown above and subsequently use it in order to create an equivalent array of cameras for the scene-side light field. Zhang et al. \cite{zhang2018generic} follow a similar approach by first relating the de-multiplexed light field in form of sub-aperture images with the scene-side, metric light field, which they base all further calculations on. 
In both works, the interpretation of the intrinsic matrix parameters is not relevant, nor are these parameters used to directly reconstruct the main lens properties. Accordingly, the presented methods do not require a re-formulation using a more complex main lens model.

Despite this fact, it is important to point out the inaccuracy in \cite{dansereau2013decoding} since several related works make only use of the proposed decoding process and assume receiving a two plane parametrization with a plane distance of $\dMm$ instead of $\dMm-\X$ as exemplarily shown in the following section.

\subsection{Depth Reconstruction}\label{subsec:PertuzAndCo}
One such case is presented by Pertuz et al. \cite{pertuz2018focus} and repeated in the follow-up work by Van Duong et al. \cite{van2019focal}. In this work a model relating the sub-aperture image shift to the object distance is deduced which translates into our notation (compare \cref{append:pertuznotation}) as
\begin{equation}\label{eq:pertuz}
	\oM(\rho)=\oM_f\left(\frac{1-a_0\cdot \rho}{1-a_1\cdot\rho}\right)
\end{equation}
with system-dependent parameters
\begin{equation}
	a_0=\frac{\fm\cdot\dML}{\px\cdot\dMm} \quad\text{and}\quad a_1=\frac{\oM_f\cdot a_0}{\fM}
\end{equation}
and a shift parameter 
\begin{equation}
	\rho=\frac{\px\cdot(\F-\Fnew)}{\fm\cdot \dML}.
\end{equation} 
There are two problems in the deduction of this model. 
First, the shift parameter $\rho$ is not correctly deduced under the premise of light field data decoded by the method of Dansereau et al. \cite{dansereau2013decoding} and second, the exit pupil is ignored. 
In the following, a corrected version is presented, which also explains, how these two problems nearly neutralize each other and lead to the same general model, albeit with different parameter interpretations.

First, the incorrect shift parameter $\rho$ is examined. In \cite{pertuz2018focus}, this parameter is described as the pixel disparity between neighboring sub-aperture images gained by the decoding process of Dansereau et al. \cite{dansereau2013decoding} and should therefore correspond to our shift parameter $\shiftwrong$ which also ignores the exit pupil. However, due to different conventions, $\rho$ is positive when focusing to a distance larger than the focus distance, whereas $\shiftwrong$ is negative in that case (compare \cref{eq:shift_alpha} for $\alpha < 1$). Accordingly, 
$\rho$ should equal $-\shiftwrong$, but transforming these parameters into a common notation leads to
\begin{align}
	-\shiftwrong(\oM) &=
	 \deltwrong\cdot\frac{\oM(\fM-\dMm)+\fM\cdot\dMm }{-\oM\cdot \fM}\nonumber\\
	 &= \frac{\px\cdot\dMm}{\fm\cdot\dML}\cdot\frac{\oM(\fM-\dMm)+\fM\cdot\dMm }{-\oM\cdot \fM}\nonumber\\
	 &\neq \frac{\px\cdot\dMm}{\fm\cdot\dML}\cdot\frac{\oM(\fM-\dMm)+\fM\cdot\dMm }{\dMm(\fM - \oM)}\\
	 &= \frac{\px\cdot(\dMm-\frac{\oM\cdot\fM}{\oM-\fM})}{\fm\cdot \dML}\nonumber\\
	 &=\frac{\px\cdot(\F-\Fnew)}{\fm\cdot \dML}=\rho.\nonumber
\end{align}

The reason for this discrepancy is the implicitly used incorrect assumption in \cite{pertuz2018focus}, that the grid of microlens image centers equals the grid of microlens centers, i.e. $\dML=\hMLI$. While a light field in general could be reparametrized 
with this step size in the $ST$ plane, this requires exact knowledge of the camera and MLA geometry, which is usually unknown and not considered in the decoding process of \cite{dansereau2013decoding}. 
Using the correct shift parameter for light field data following \cite{dansereau2013decoding}  instead and rearranging the corresponding \cref{eq:o_incorrect} results in
\begin{align}
	\oMwrong(\shiftwrong)&= \frac{\fM\cdot\dMm\cdot\deltwrong}{\shiftwrong\cdot\fM-\deltwrong(\fM-\dMm)}\nonumber\\
	&=\oM_f\left(\frac{(\dMm-\fM)\cdot\deltwrong}{\shiftwrong\cdot\fM - \deltwrong(\fM-\dMm)}\right)\nonumber\\
	&=\oM_f\left(\frac{1}{1+\left(\frac{\fM}{\deltwrong(\dMm-\fM)}\right)\cdot \shiftwrong}\right)=\oM_f\left(\frac{1}{1-\left(\frac{\oM_f\cdot\fm\cdot\dML}{\dMm^2\cdot\px}\right)\cdot (-\shiftwrong)}\right).
\end{align}
This correction of \cref{eq:pertuz} still ignores the exit pupil just as \cite{pertuz2018focus}, but is considerably simpler than the original model since only a single system-dependent parameter $\frac{\oM_f\cdot\fm\cdot\dML}{\dMm^2\cdot\px}$ is present compared to the two parameters $a_0$ and $a_1$ in \cref{eq:pertuz}.

Finally, by introducing a non-zero distance $\X$ and thereby using $\oM$ and $\shift$ instead of $\oMwrong$ and $\shiftwrong$, the full model can be deduced from \cref{eq:o_correct} as
\begin{align}
	\oM(\shift)&= \frac{\fM\cdot(\dMm\cdot\delt-\shift\cdot\X)}{\shift\cdot(\fM-\X)-\delt(\fM-\dMm)}\nonumber\\
	&=\oM_f\left(\frac{(\dMm-\fM)\cdot(\fM\cdot\dMm\cdot\Delta-\fM\cdot\X\cdot (-\shift))}{(\fM\cdot\dMm)\cdot(\Delta(\dMm-\fM)-(\X-\fM)\cdot\shift)}\right)\nonumber\\
	&=\oM_f\left(\frac{\fM\cdot\dMm\cdot\Delta-\fM\cdot\X\cdot \shift}{\fM\cdot\dMm\cdot(\Delta-\frac{\X-\fM}{\dMm-\fM}\cdot\shift)}\right)\nonumber\\
	&=\oM_f\left(\frac{1-\left(\frac{\X}{\Delta\cdot\dMm}\right)\cdot\shift}{1-\left(\frac{\X-\fM}{\Delta\cdot(\dMm-\fM)}\right)\cdot \shift}\right)
\end{align}
and to align this model with the inverted shift direction of \cite{pertuz2018focus} we define
\begin{align}\label{eq:pertuz_corrected}
\oM_f\left(\frac{1-\left(\frac{\X}{\Delta\cdot\dMm}\right)\cdot\shift}{1-\left(\frac{\X-\fM}{\Delta\cdot(\dMm-\fM)}\right)\cdot \shift}\right)
=\oM_f\left(\frac{1-\left(\frac{-\X}{\Delta\cdot\dMm}\right)\cdot(-\shift)}{1-\left(\frac{\fM-\X}{\Delta\cdot(\dMm-\fM)}\right)\cdot (-\shift)}\right)=: \oM_f\left(\frac{1-a_0\cdot (-\shift)}{1-a_1\cdot (-\shift)}\right).
\end{align}
This model has the same general form as \cref{eq:pertuz} as proposed by Pertuz et al. \cite{pertuz2018focus}, which explains the reasonable experimental results in that work. Nevertheless, the interpretation of the system-dependent parameters $a_0$ and $a_1$ as given in \cite{pertuz2018focus} and repeated in \cite{van2019focal} is incorrect, which is also verified in experiment (V) in the following section. This different interpretation could lead to problems, when the model needs to be fitted to data and the initial parameter values are based on the incorrect direct calculation. 

Overall, the results of Pertuz et al. \cite{pertuz2018focus} are not entirely incorrect, but the used light field representation based on \cite{dansereau2013decoding} is simply not matching the implicit assumptions used for the model deduction. More specifically, the main problem of \cite{pertuz2018focus} is the definition of the shift parameter $\rho$ for light field data decoded similar to \cite{dansereau2013decoding}, but based on known microlens centers instead of MICs. While the light field could theoretically be reparametrized 
by using the parallel projections of the microlens centers onto the sensor, this would require exact knowledge of the SPC intrinsics, namely the MLA parameters as well as its placement relative to the main lens and sensor.

\section{Evaluation}\label{sec:Evaluation}
\subsection{Simulation Environment}\label{subsec:Simulation}
\begin{figure}[h!]
	\centering
	\begin{tikzpicture}
		\draw[fill=white] (0,0) rectangle ++(0.405\textwidth,0.225\textwidth);
		\node[inner sep=5pt, outer sep=0, anchor=south west] (fig0) at (0,0) {\includegraphics[width=0.4\textwidth]{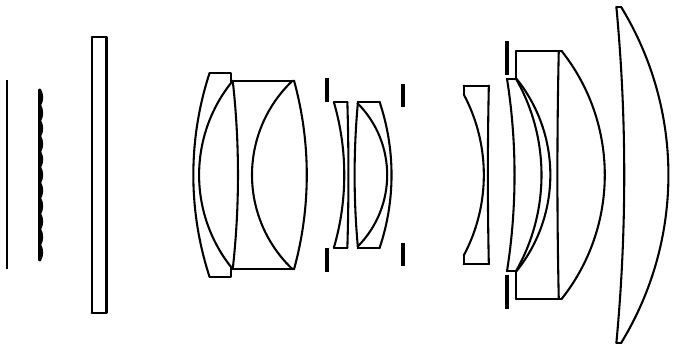}};

		\node[inner sep=0pt, outer sep=12pt, anchor=west] (fig1) at (fig0.east) {\includegraphics[width=0.4\textwidth]{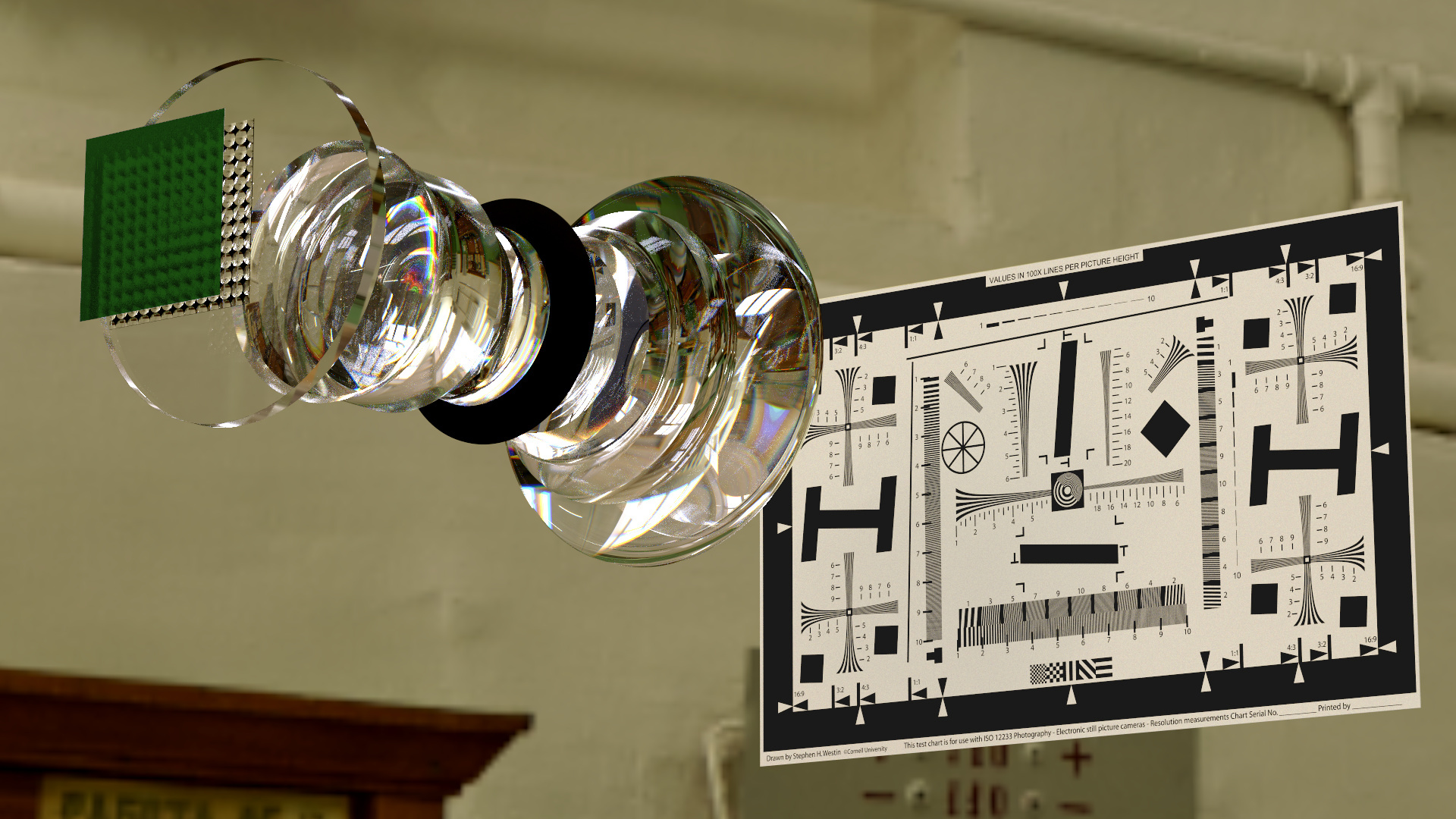}};
	\end{tikzpicture}
	\caption[Cross-section]{Cross-section and rendering of an exemplary evaluation setup: A fully modeled lens is combined with a two plane MLA model and a sensor in order simulate a plenoptic camera via ray tracing. Calibration patterns are placed at different distances in front of this setup to verify the analytical models. Note, that the housing of camera and lens were removed for the purpose of visualization and the ISO 12233 pattern\footnotemark{} is used with permission of Cornell University.}
	\label{fig:eval_rendering}
\end{figure}

Since SPCs with exchangeable lenses are not commercially available at the time of writing and custom-built solutions are costly as well as prone to misalignment of the optical components, we resort to synthetic experiments via an extension of the ray-tracing solution we provided in \cite{michels2018simulation}. Our publicly available\footnote{\url{https://gitlab.com/ungetym/blender-camera-generator}}, updated version of the Blender \cite{blender} Add-On expands the original simulation by the following aspects:
\begin{itemize}
	\item Simulation of aspherical 
	 lenses and zoom lenses
	\item Configurable MLA pose, thickness and IOR
	\item Automatic focusing with lens group movement based on paraxial approximations
	\item Integration of Claff's lens collection \cite{claff} and a collection of sensor presets
	\item Assisted plenoptic camera (SPC and FPC) configuration based on the ideas of \cite{michels2021ray}
\end{itemize}
This simulation facilitates a quick generation of a broad range of plenoptic cameras, such as the one exemplarily shown in \cref{fig:eval_rendering}, and is used in the following to validate the formal analysis of the sections \ref{subsec:refocus}, \ref{sec:erroranalysis} and \ref{subsec:PertuzAndCo}.
\footnotetext{\url{https://www.graphics.cornell.edu/~westin/misc/res-chart.html}}

\subsection{Experiments}\label{subsec:Experiments}
For the validation the five lenses listed in \cref{table:lenses} were selected from the database \cite{claff}. 
\begin{table}[h]
	\centering
	\setlength{\belowcaptionskip}{4pt}
	\resizebox{\textwidth}{!}{%
		\begin{tabular}{|Sc||Sc|Sc|Sc|Sc||Sc|Sc|Sc|}
			\hline
			& \multicolumn{4}{c||}{ Finite Focus} & \multicolumn{3}{c|}{ Infinite Focus}\\
			\cline{2-8}
			Lens Model& $\fM$ (mm)& $\X$ (mm) & $\frac{\X}{\fM}$ & \makecell{ Focus \\ Dist. (mm)}& $\fM$ (mm)& $\X$ (mm) & $\frac{\X}{\fM}$\\
			\hline
			\hline
			\makecell{ Rodenstock \\ Sironar-N 100mm f5.6}& 99.998 & 0.194 & 0.002 & 500.0&99.998&0.194 & 0.002\\
			\hline
			\makecell{ Zeiss \\ Batis 1.8-85} & 82.047 & 40.652 & 0.496 & 500.0&82.860&39.573&0.478\\
			\hline
			\makecell{ Ricoh smc Pentax-A \\ 200mm F4 Macro ED}& 167.994 & 99.908 & 0.595 & 500.0&173.115&91.854&0.531\\
			\hline
			\makecell{ Canon \\ EF85mm f1.8 USM}& 84.998 & -28.938 & -0.341 & 300.0& 84.998 & -28.938 & -0.341\\
			\hline
			\makecell{Olympus Zuiko \\ Auto-Zoom 85-250mm F5}& 85.120 & -60.219 & -0.708 & 300.0&85.004&-60.308&-0.709\\
			\hline
	\end{tabular}}
	\caption{Overview of the simulated main lenses and their properties in the finite and infinite focus setup. Note, that the focal length of a lens can vary in different setups due to lens group movements involved in refocus or zoom operations.\label{table:lenses}}
\end{table}
While the first lens presents the ideal case of $\X\approx0 mm$, i.e. the exit pupil coinciding with the camera-side principal plane, the remaining four lenses present interesting cases with varying relationships between $\X$ and the lenses' focal lengths $\fM$.
Each of these lenses is used in two SPC configurations, one with a finite focus distance $\oM_f<\infty$ and another one focused at infinity, $\oM_f=\infty$. To this end, the MLA placement with respect to the main lens, i.e. the distance $\dMm$ between the MLA and the camera-side principal plane of the main lens, can be calculated by the thin lens \cref{eq:thinlens} as $\dMm = \frac{\fM\cdot \oM_f}{\oM_f-\fM}$ for the finite case and simply be set to $\dMm = \fM$ for $\oM_f=\infty$. 
The remaining microlens parameters are automatically initialized using the main lens and sensor properties to fulfill the following two constraints. 
First, the microlens f-number needs to match that of the main lens in order to optimally cover the sensor area \cite{ng2005lightfieldcamera} and second, a predefined number of $129\times 129$ microlens images should be visible on the sensor in order to guarantee this resolution for the sub-aperture images. 
The resulting parameters are then fine-tuned by hand to accommodate the approximating 
nature of the f-number constraint and to guarantee that the MICs coincide with the centers of sensor pixels. 
This optimal MIC positioning has two effects: First, it renders the resampling during the decoding process of \cite{dansereau2013decoding} unnecessary. 
After normalizing the raw images with white images to get rid of vignetting effects \cite{yu2004practical}, the sub-aperture images in our setups can directly be extracted by combining the same relative sensor pixels from each microlens image \cite{ng2005lightfieldcamera}. 
And second, the evaluation can concentrate on validating the refocusing itself instead of additionally dealing with the compensation of interpolation artifacts from the decoding process. 
\noindent The full setups can be found in \cref{append:setups} and with each of these ten setups, five experiments are performed:

\noindent\textbf{(I)} \textbf{MICs and Exit Pupil:} The exit pupil as origin of the MICs is verified by first tracing ray bundles from the main lens aperture center through the main lens and MLA onto the sensor. This results in a set of sensor hits for every microlens. 
Due to the small variance within such a set, the mean is considered to represent the ground truth 
position of the microlens image center. 
In a second step, rays are traced from these sensor positions through the corresponding microlens centers and the convergence location of the resulting ray bundle is calculated in two ways. 
First by performing a line search along the optical axis for the minimum blur spot position of the ray bundle.
\begin{figure}[h!]
	\centering
	\includegraphics[page=1]{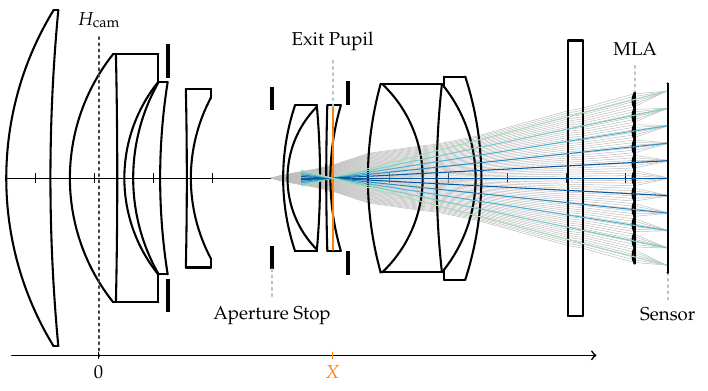}
	\caption{Experiment (I): The two steps of the MIC/exit pupil verification visualized for the Zeiss Batis 1.8-85. Rays (light gray) are traced from the main lens aperture center through the main lens and MLA onto the sensor. The resulting means of the sensor hits per microlens represent the MICs. The exit pupil as the approximate source of these points is verified by backward tracing rays (blue/green) from the MICs through the respective microlenses and calculating the minimum blur spot position as well as the mean and variance of intersections along the optical axis.}
	\label{fig:MIC_verification}
\end{figure}
And second based on the rays' intersections with the optical axis. For these intersections the mean and variance are calculated and presented alongside the minimum blur spot position.	 
This whole process is also visualized in \cref{fig:MIC_verification}.

\noindent\textbf{(II)} \textbf{$\oM$ and $\shift$:} A calibration pattern, more specific a Siemens star with four spokes, is placed at various distances in front of the camera and after demultiplexing 
the plenoptic camera image, the sub-aperture image shift $\shift$, which is required to focus onto the given target distance $\oM$, is measured. This is done via line search, i.e. by repeatedly refocusing the image with a simple shift-and-sum algorithm \cite{ng2005lightfieldcamera} and calculating the sharpness of the refocused image.
Here, the variance of the Laplacian \cite{pertuz2013analysis} is used as metric for image sharpness and the shift value with the highest score is considered the optimum. 
This procedure results in tuples of ground truth distances $\oM$ and measured shifts $\shift$, which are then used to verify the connection $\shift(\oM)$ as formally described in \cref{eq:S_correct}. 

\noindent To validate the inverted connection $\oM(\shift)$ in \cref{eq:o_correct}, all images are refocused for a given set of shift values. For each of these shift values, the object distance associated with the best focused image is considered the measured object distance for the respective shift value. The resulting tuples of preset shifts and measured distances are then used to verify $\oM(\shift)$.
	
\noindent\textbf{(III)} \textbf{$\errorowrong$ Validation:} The data of experiment (II) is further used in order to verify the error model $\errorowrong$. In detail, the measured shift $\shift$ for a known target distance $\oM$ is used with \cref{eq:o_incorrect} to approximate $\oMwrong(\shift)$ and calculate the measured relative error according to \cref{eq:error_o}. This error is then compared to the expected error gained by directly calculating $\shift$ based on the camera's properties instead of measuring it. 
	
\noindent\textbf{(IV)} \textbf{$\errorswrong$ Validation:} Moreover, the images of (II) are also used to verify the error model $\errorswrong$ presented in \cref{sec:erroranalysis}.
First, for every target distance $\oM$ the incorrect shift $\shiftwrong(\oM)$ is calculated based on the assumption $\X = 0$ as in \cref{eq:S_incorrect} and the images of the patterns at different positions are all refocused with this parameter.
The target distance corresponding to the sharpest of the refocused images approximates $\oM(\shiftwrong)$ and is then used to measure $\errorswrong$ as in \cref{eq:error_o2}. Again, the measured values are compared to the errors gained by directly calculating $\oM(\shiftwrong)$.

\noindent Note, that instead of verifying $\errorowrong$ and $\errorswrong$, the equations \ref{eq:S_incorrect} and \ref{eq:o_incorrect} as well as the shift error model in \cref{eq:error_S} could also directly be validated with the data measured and calculated in the experiments (III) and (IV). However, these equations do not include a comparison between the incorrect estimations and the ground truth refocusing distances. Therefore, the indirect validation of those models by means of the resulting refocusing errors was preferred.
	
\noindent\textbf{(V)} \textbf{Validation of \cref{subsec:PertuzAndCo}:} As analyzed in that section, the overall formulation of the model presented by Pertuz et al. \cite{pertuz2018focus} is correct, but the model parameters have a different interpretation under the assumption of light field data decoded by the method of \cite{dansereau2013decoding}.
To verify the corrected model, the parameters $a_0$ and $a_1$  are first calculated based on the formula of Pertuz \cite{pertuz2018focus}, then according to our model from \cref{subsec:PertuzAndCo}, and finally fitted to the data of (II), i.e. the set of shift-distance-pairs with measured shifts and ground truth target distances.
For this parameter fitting, a grid search for the best parameters by means of the RMSE was performed with a grid explicitly containing both directly calculated parameter sets.

\subsection{Results and Discussion}\label{subsec:Results}
\begin{figure}[h]
	\centering
	\includegraphics[page=1]{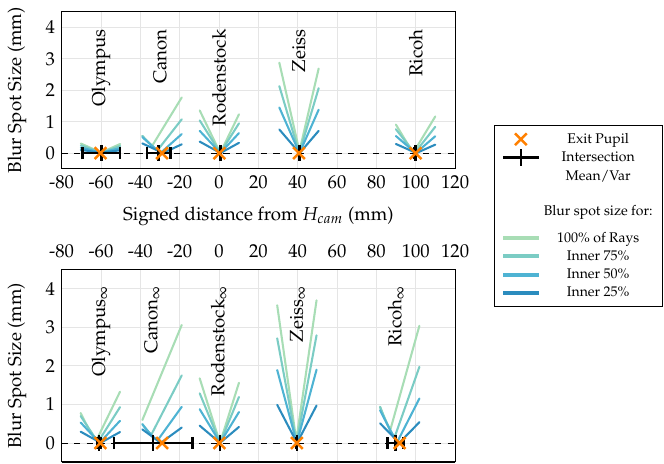}
	\caption{Results of experiment (I). The orange markers indicate the exit pupils' locations on the optical axis (horizontal, dotted line) with respect to the respective principal planes $\Hcam$. The black markers show the mean and variance of the intersection points between the optical axis and the rays traced back from the MICs through the microlens centers. The colored functions show the blur spot sizes for different subsets of these ray bundles close to the exit pupil (compare \cref{fig:MIC_verification}). These subsets contain the respective portion of rays from the bundle, which are closest to the optical axis.}
	\label{fig:results_I}
\end{figure}
\noindent\textbf{(I)} As shown in \cref{fig:results_I}, the ray bundle consisting of rays running from the calculated ground truth MICs through the microlens centers in general converge towards the exit pupil in all setups.
The closer the ray origins are to the optical axis, compare e.g. the inner 25\% of rays shown in \cref{fig:results_I}, the closer the minimum blur spot is located to the exit pupil. The deviations for larger sets of rays including the outer most MICs, as especially visible for the Canon setups, can be explained by the dependency of the exit pupil's location on the viewing angle. Similar to the curved focal plane in lenses with a significant non-zero Petzval field curvature, the exit pupil cannot be well approximated by a plane in some setups and thereby affects the MIC grid on the sensor in a non-linear fashion \cite{smith2008modern}. This is a clear limitation of our model, which is built upon paraxial approximations. 
Nevertheless, considering that even in extreme cases like the outer-most microlenses in the Canon setups, the origin of the MICs is located close to the exit pupil plane, these results again confirm the observation of Hahne et al. \cite{hahne2018baseline} and thus justify the recommendation to examine the necessity of including the exit pupil into the lens model. 

\begin{figure}[h]
	\centering
	\includegraphics[page=1]{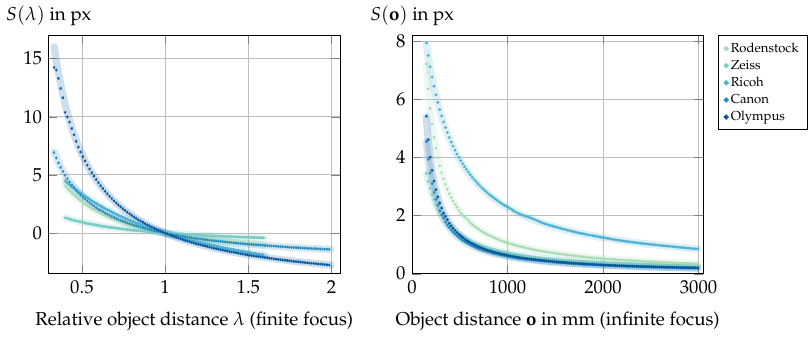}
	\caption{Results of experiment (II): Verification of the model $\shift(\oM)$ according to \cref{eq:S_correct}. The data points represent measured shift values for the respective object distances and the underlying lines represent the expected, directly calculated values. Left: Error for the setups with finite focus based on the relative object distance $\frac{\oM}{\oM_f}$. Right: Error for setups focused at infinity.}
	\label{fig:results_model_s}
\end{figure}

\noindent\textbf{(II)} The connection between $\oM$ and $\shift$ as described by \cref{eq:S_correct} is verified by the measurements presented in \cref{fig:results_model_s}.
The mean of the absolute differences $|\shift(\oM)_{measured}-\shift(\oM)_{GT}|$ between the measured and directly calculated ground truth shift values across all 10 setups is 0.008 px with a variance of 0.0026 px. The worst single setup is the finitely focused Olympus setup with a mean of absolute differences of 0.04 px and a variance of 0.025 px. These values are in the range of expected inaccuracies resulting from the image processing methods involved. Especially the interpolation steps required by the shift-and-sum refocusing \cite{ng2005lightfieldcamera}, but also the rather simple (de)focus measure, which is prone to interference errors, are limiting factors, which prevent a higher accuracy.
\begin{figure}[h]
	\centering
	\includegraphics[page=1]{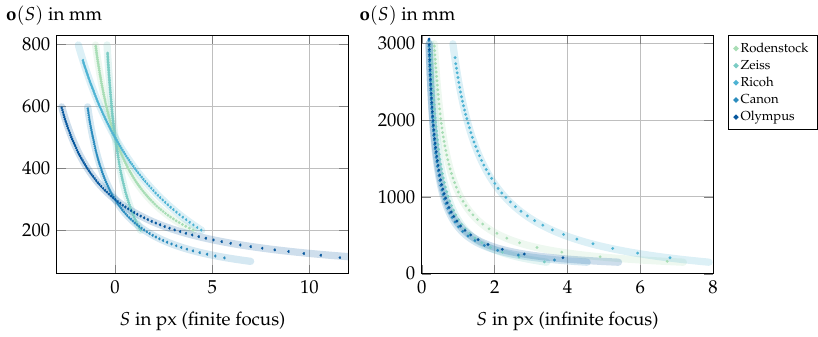}
	\caption{Results of experiment (II): Verification of the model $\oM(\shift))$ according to \cref{eq:o_correct}. The data points represent measured focus distance values for the respective shifts and the underlying lines represent the expected, directly calculated values.}
	\label{fig:results_model_o}
\end{figure}

\noindent A similar situation is observable in the inverse case, i.e. for the model $\oM(\shift)$ from \cref{eq:o_correct}, as presented in \cref{fig:results_model_o}. Due to the wide range of target distances, in this case the relative absolute differences $\frac{|\oM(\shift)_{measured}-\oM(\shift)_{GT}|}{\oM(\shift)_{GT}}$ 
are calculated to quantify the results from \cref{fig:results_model_o}. 
For the finite setups, the mean of these relative absolute differences is 0.13 \% with a variance of $2\cdot10^{-5}$ \%, while in the infinite cases, the overall mean is 0.72 \% with a variance of 0.01 \%. Here, the infinitely focused Zeiss setup has the worst performance with a relative absolute difference mean of 0.9 \% and a variance of 0.004 \%. The worse performance of the infinitely focused setups is a consequence of the smaller range of shift values representing a larger refocusing range (compare \cref{fig:results_model_o}) resulting in a greater susceptibility to small shift changes. Nevertheless, the overall performance confirms the model $\oM(\shift)$, again within the constraints posed by the involved image processing steps.

\begin{figure}[h]
	\centering
	\includegraphics[page=1]{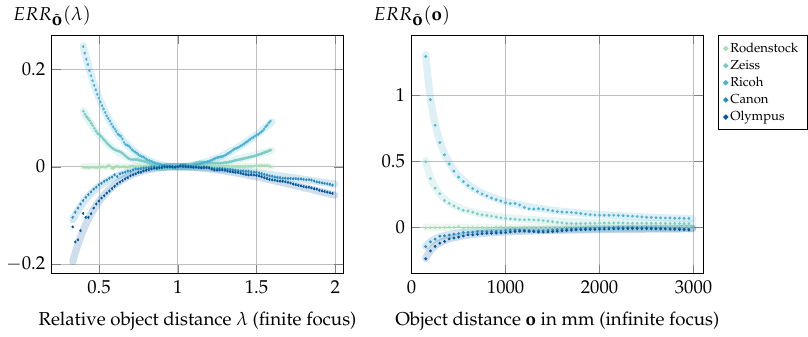}
	\caption{Results of experiment (III): Relative object distance error resulting from a correctly estimated shift, but incorrect object distance estimation based on the assumption $\X=0$. 
	The thick lines indicate the predicted ground truth error based on \cref{eq:error_o} and \cref{eq:error_o_lambda} and the points indicate the measurements. Left: Error for the setups with finite focus. For a comparative visualization of the different target distance ranges, the results are shown based on the relative object distance $\lambda=\frac{\oM}{\oM_f}$. Right: Error for setups focused at infinity.}
	\label{fig:results_error}
\end{figure}
\noindent\textbf{(III)}
The results of the $\errorowrong$ verification experiment are shown in \cref{fig:results_error}. These graphs again show the error based on measurements compared to the directly calculated ground truth error based on the error models presented in \cref{sec:erroranalysis}. 
The mean of absolute differences between the measured and calculated error values is 0.003 with a variance of $10^{-5}$, which validates our models limited only by the accuracy of image processing methods and the optical properties of the chosen lenses. More specific, the small deviations from the expected values, which are even present in the baseline case for the Rodenstock lens, can be explained as in experiment (II) by the interpolation operations required by the shift-and-sum refocusing algorithm \cite{ng2005lightfieldcamera} for non-integer shift values. The larger deviation visible at close range for the Olympus lens with finite focus distance is a result of the optical limits of this lens which can be explained as follows. A single sub-aperture image contains one pixel per microlens. Hence, such an image can be considered sharp, if the calibration target is imaged onto the MLA with all blur spot sizes smaller than the microlens diameter $\dML$. For the Olympus lens however, these blur spot sizes increase more drastically in the close range than those of the other lenses, leading to severe defocus blur even in the sub-aperture images. This in turn can produce interference artifacts during the refocusing and thereby lead to fluctuating contrast measurements. And since the refocus distance $\oM$ is determined by these measurements, this affects the error calculated by $\frac{\oMwrong(S(\oM))-\oM}{\oM}$ as in \cref{eq:error_o}.

\begin{figure}[h]
	\centering
	\includegraphics[page=1]{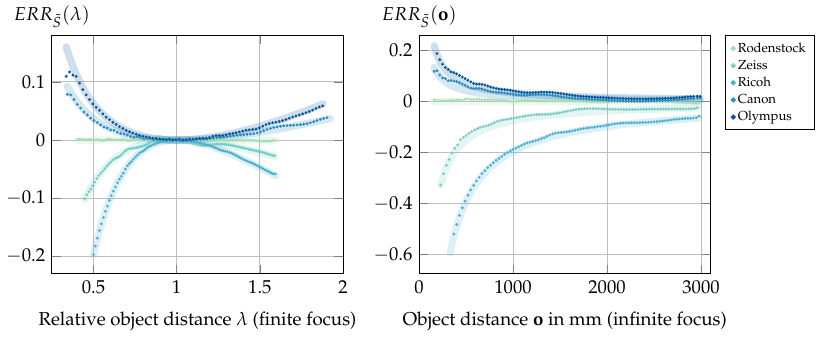}
	\caption{Results of experiment (IV): Relative object distance error resulting from an incorrectly calculated shift ignoring the exit pupil, combined with a correct object distance estimator. The thick lines indicate the predicted ground truth error based on \cref{eq:error_o2} and \cref{eq:error_o2_lambda} and the points indicate the measurements. Left: Error for the setups with finite focus. For a comparative visualization the results are shown based on the relative object distance $\frac{\oM}{\oM_f}$, i.e. the finite focus distance for the respective setups is met at $\frac{\oM}{\oM_f}=1$. Right: Error for setups focused at infinity.}
	\label{fig:results_error_inf}
\end{figure}

\noindent\textbf{(IV)}
While experiment (III) validated the error model in the case of a correctly estimated shift combined with an oversimplified distance estimator, \cref{fig:results_error_inf} shows, that the inverse problem of an incorrectly calculated shift used with the shift-and-sum refocusing is also modeled correctly. The mean of absolute differences between the measured and directly calculated error values is 0.004 with a variance of $2.47\cdot 10^{-5}$. 

Overall, the experiments (III) and (IV) confirm the formal deductions of \cref{sec:erroranalysis} and thus again justify the warning to mind the exit pupil when modeling a standard plenoptic camera. However, the results also hint at further minor optical or algorithmic aspects not being accounted for. 
In all cases the mean of the absolute differences between measured values and ground truth model is up to two orders of magnitude larger than the respective variance.
This is the result of a nearly constant over- or underestimation and could indicate a systematic error, which can e.g. be caused by the refocusing model being limited to paraxial calculations.

\begin{table}[h]
	\centering
	
	\renewcommand{\arraystretch}{1.3}
	\setlength{\belowcaptionskip}{6pt} 
	
	\resizebox{\textwidth}{!}{%
		\begin{filecontents*}{data/verify_model.csv}
Setup,a012,a112,RMSE12,a0our,a1our,RMSEour,a0,a1,RMSE
Rodenstock,0.092658,0.4633,21.03,-0.00014433,0.3705,0.63755,-0.00038363,0.36951,0.48873
Zeiss,0.18533,1.1294,28.048,-0.13102,0.81304,1.5106,-0.13109,0.81361,1.5053
Ricoh,0.11392,0.33906,89.771,-0.074345,0.15079,0.52064,-0.074032,0.15111,0.50816
Canon,0.13411,0.47334,45.504,0.026304,0.36553,0.73644,0.025547,0.36568,0.58973
Olympus,0.067415,0.2376,37.489,0.022672,0.19286,1.1189,0.0213,0.19146,0.62405
Rodenstock$_\infty$,0.092646,55.588,96.871,-0.00018016,55.495,7.1889,0.0019845,55.625,6.8321
Zeiss$_\infty$,0.10697,322.73,153.3,-0.097725,322.53,9.6725,-0.085743,323.61,8.3512
Ricoh$_\infty$,0.063496,366.78,364.78,-0.071746,366.65,6.7611,-0.068866,367.54,5.0255
Canon$_\infty$,0.14578,1715.1,67.397,0.037024,1715,10.261,0.045757,1724.1,8.6856
Olympus$_\infty$,0.13697,96.683,46.52,0.056801,96.603,8.2331,0.06014,96.631,8.074
\end{filecontents*}
\pgfplotstabletypeset[
		col sep=comma,
		string type,
		every head row/.style={
			before row={
				\hline
				& \multicolumn{3}{c|}{Pertuz et al.\cite{pertuz2018focus}} & \multicolumn{3}{c|}{Ours (\cref{eq:pertuz_corrected})} & \multicolumn{3}{c|}{Fitted} \\
				\cline{2-10}
			},
			after row=\hline\hline,
		},
		every last row/.style={after row=\hline},
		display columns/0/.style={column name=Setup, string type, column type={|c}},
		display columns/1/.style={column name=$a_0$, column type={|c}},
		display columns/2/.style={column name=$a_1$},
		display columns/3/.style={column name=RMSE, column type={c|}},
		display columns/4/.style={column name=$a_0$},
		display columns/5/.style={column name=$a_1$},
		display columns/6/.style={column name=RMSE, column type={c|}},
		display columns/7/.style={column name=$a_0$},
		display columns/8/.style={column name=$a_1$},
		display columns/9/.style={column name=RMSE, column type={c|}},
		]{data/verify_model.csv}
	}
	\caption{Results of experiment (V): Model parameters and the resulting RMSE in mm for the 10 SPC setups. The parameters for our method and Pertuz et al. \cite{pertuz2018focus} are directly calculated based on the known optical properties. The fitted parameter sets were acquired via grid search and subsequent local optimization to fit the model to the data from experiment (II).}
	\label{table:results_model}
\end{table}
\noindent\textbf{(V)} Regarding the correction of the model from \cite{pertuz2018focus} 
in \cref{subsec:PertuzAndCo}, the results as presented in \cref{table:results_model} show, that the corrected model appropriately describes the connection between refocus distance and the sub-aperture image shift. In all setups the RMSE of our directly calculated model is within 1.75 mm of the RMSE of the fitted model. In addition, the fitted parameters are well approximated by the direct calculation with our model. On the other hand, the model of \cite{pertuz2018focus} can be regarded as incorrect for this light field parametrization with one exception - the parameter $a_1$ in the setups focused at infinity is in general correct which can be explained as follows. Neither the original model of \cite{pertuz2018focus} nor our correction directly considers the case $\oM_f=\infty$, instead a large focus distance of $\oM_f=10^6 m$ is used as approximation in these cases. For such a focus distance, the distance $\dMm$ between main lens and the MLA is close to the focal distance of the main lens, i.e. $\dMm=\fM+\epsilon$ for some small value $\epsilon>0$. With this formulation, the correctness of $a_1$ in the infinite cases can formally be explained by 
\begin{align}
\frac{a_{1,\text{Pertuz}}}{a_{1,\text{our}}}&=\left(\frac{\oM_f\cdot a_{0,\text{Pertuz}}}{\fM}\right) \cdot \left(\frac{\Delta(\dMm-\fM)}{\fM-\X}\right) =\frac{\oM_f\cdot \frac{\fm\cdot\dML}{\px\cdot\dMm}}{\fM}\cdot \frac{\frac{\px \cdot(\dMm-\X)}{\fm\cdot \dML}\cdot(\dMm-\fM)}{\fM-\X}\nonumber\\
&=\oM_f\cdot \underbrace{\frac{\dMm-\fM}{\dMm\fM}}_{=1/\oM_f}\cdot\frac{\dMm-\X}{\fM-\X}= \frac{\fM + \epsilon-\X}{\fM-\X}=1+\frac{\epsilon}{\fM-\X}\approx 1.
\end{align}

\section{Conclusion and Limitations}\label{sec:Conclusion}
Overall, this work shows, that the exit pupil can play a crucial role when modeling the relation between the camera-side and scene-side light fields. The connection between sub-aperture image shift and refocusing distance is derived analogously to previous work, but under consideration of an exit pupil not coinciding with the principal plane of the main lens. Based on this deduction, two error models for the relative refocus distance are created and validated. The subsequent review of previous work shows, that a sufficiently general formulation of the SPC calibration model in most methods absorbs these errors, albeit leading to an incorrect interpretation of the model parameters. Exemplarily a correction of the work of Pertuz et al. \cite{pertuz2018focus} is presented and validated.

Nevertheless, despite the good evaluation results, there are several limitations to this work. First of all, the experiments are performed on simulated data. While the ray-tracing based lens simulation has been verified to exhibit the optical properties stated in the respective lens patents, i.e. aberration and distortion measurement results from the patents could be reproduced, there still is a gap between simulation and reality. On one hand, the specified lens parameters could differ from the final production lens due to manufacturing inaccuracies or even deliberate parameter obfuscation by the lens manufacturer to hide specific lens details. On the other hand, the used framework given by Blender \cite{blender} does not include wave optic effects such as diffraction \cite{freniere1999edge}\cite{mahan2018monte}. Without these effects the simulated optics are not diffraction-limited and therefore might produce images sharper than their real pendants.

\noindent Further limitations are concerned with the formal lens model used for the error deduction. First, the microlenses in our model are still formally regarded as thin lenses. While Hahne et al. \cite{hahne2016refocusing} use a thick lens model with explicit microlens principal planes, it was decided to leave this aspect out of the theoretical discussion for reasons of clarity and comprehensibility. However, the microlens thickness was factored in while performing the experiments and \cref{append:thick} shows, that an extended model does not change the equations deduced in \cref{sec:SPC_Optics} and \cref{sec:erroranalysis}. 

\noindent Another limitation, which does affect the validity of these equations, is the restriction to paraxial models, more specific the repeated use of the thin lens equation \ref{eq:thinlens} in various calculations as well as fixed positions for the principal planes and exit pupils. The thin lens equation describes the relation between the object distance, image distance and a lens' focal length and is usually only valid along the optical axis. With growing distance from this axis, third-order aberrations like the Petzval field curvature, i.e. a curved focus surface, might affect the refocusing distance \cite{smith2008modern}. Furthermore, as seen for the evaluated Canon lens in experiment (I), the position of the exit pupil might also vary depending on the viewing angle and thereby reduce the applicability of the deduced models.

Further work on the listed limitations is not expected to significantly alter the presented results, as these are already within the expected accuracy bounds set by the involved image processing steps. Instead, future work could target the second type of plenoptic cameras, namely the FPC, for which a multitude of different calibration methods exists, that also differ in the assumed lens models and could benefit from minding the gap between principal plane and exit pupil.

\authorcontributions{Conceptualization, T.M.; methodology, T.M.; software, T.M.; validation, T.M.; formal analysis, T.M. and D.M.; investigation, T.M. and D.M.; resources, T.M.; data curation, T.M.; writing---original draft preparation, T.M.; writing---review and editing, T.M., D.M. and R.K.; visualization, T.M.; supervision, R.K.; project administration, R.K. and T.M.; funding acquisition, R.K. All authors have read and agreed to the published version of the manuscript.}

\funding{This research received no external funding.}

\dataavailability{Data available in two publicly accessible repositories
	The data presented in this study is openly available at \url{https://gitlab.com/ungetym/SPC-revisited} and the camera simulation is available at \url{https://gitlab.com/ungetym/blender-camera-generator}.} 



\acknowledgments{We would like to express our thanks to William 'Bill' Claff, who supported us with his lens database at \url{https://www.photonstophotos.net}.}

\conflictsofinterest{The authors declare no conflict of interest. The funders had no role in the design of the study; in the collection, analyses, or interpretation of data; in the writing of the manuscript; or in the decision to publish the results.} 



\abbreviations{Abbreviations}{
The following abbreviations are used in this manuscript:\\
\noindent 
\begin{tabular}{@{}ll}
FPC & Focused Plenoptic Camera\\
MI & Microlens Image\\
MIC & Microlens Image Center\\
ML & Microlens\\
MLA & Microlens Array\\
SPC & Standard Plenoptic Camera
\end{tabular}
}

\appendixstart
\appendix

\section[\appendixname~\thesection]{Notation}\label{append:notation}
\begin{table}[h]
	\centering
	\renewcommand{\arraystretch}{1.1}
	\begin{tabular}{|Sc|Sc|}
		\hline
		& Camera Parameters\\
		\hline
		\hline
		$\fM$ & Main lens focal length \\
		$\Hscene$ & Scene-side principal plane of the main lens \\
		$\Hcam$ & Camera-side principal plane of the main lens \\
		$\X$ & \makecell{Signed distance between $\Hcam$ and the exit pupil\\ measured along the optical axis} \\
		$\fm$ & Microlens focal length \\
		$\HsceneML$ & Scene-side principal plane of a microlens \\
		$\HcamML$ & Camera-side principal plane of a microlens \\
		$\oM$ & Object distance measured from $\Hscene$ \\
		$\iM$ & Image distance measured from $\Hcam$ \\
		$\oM_f$ & Focus distance\\
		$\dMm$ & Distance between the MLA's $\HsceneML$ and the main lens $\Hcam$\\
		$\dML$ & \makecell{Microlens pitch}\\
		$\px$ & Pixel pitch, i.e. edge length of a square sensor pixel\\	
		\hline
		\hline
		& Light Field Parametrization\\
		\hline
		\hline
		$\F$ & \makecell{Distance between $UV$ and $ST$ plane}\\
		$\deltaUV$ & Step size in the virtual lens plane\\
		$\deltaXY$ & Pixel pitch on the virtual sensor\\
		$\Delta$ & Scaling between the $UV$ and $ST$ step sizes, i.e. $\Delta=\deltaUV/\deltaXY$\\
		$L_\F$ & Integer indexed 4D light field\\
		$i,j,k,l$ & Integer light field coordinates\\
		$\tilde{L}_\F$ &Metric parametrization of the 4D light field\\
		$s,t,u,v$ & Metric light field coordinates\\
		\hline
		\hline
		& Refocusing and Error Analysis\\
		\hline
		\hline
		$\Fnew$ & Distance between $UV$ and $S'T'$ plane for refocusing\\
		$\alpha$ & Refocusing parameter defined by $\alpha=\F/\Fnew$\\
		$\tilde{L}_{\Fnew}$ &\makecell{$\tilde{L}_F$, but for $S'T'$ plane}\\
		$\shift$ & Sub-aperture image shift in pixels\\
		$\shift(\oM)$ & $\shift$ for a given refocusing distance $\oM$\\
		$\oM(\shift)$ & Refocusing distance $\oM$ for a given shift $\shift$\\
		$\deltwrong$ & Quotient $\Delta$ based on the assumption $\X=0$\\
		$\shiftwrong$ & \makecell{Sub-aperture image shift in pixels based on the assumption $\X=0$}\\
		$\oMwrong$ & Refocusing distance based on the assumption $\X=0$\\
		$\lambda$ & \makecell{Quotient $\oM/\oM_f$}\\
		$\errorowrong$ & \makecell{Relative refocusing distance error for incorrect target distance model}\\
		$\errorswrong$ & \makecell{Relative refocusing distance error for incorrect shift model}\\
		\hline
	\end{tabular}
\end{table}

\section[\appendixname~\thesection]{Revisited Literature - Notation Transfer}
In the following sections, the notation of \cite{hahne2016refocusing}, \cite{dansereau2013decoding} and \cite{pertuz2018focus} are transferred into the notation used in this work.

\subsection[\appendixname~\thesubsection]{Hahne et al.\cite{hahne2016refocusing}}\label{append:hahnenotation}
\begin{table}[h]
	\centering
	\renewcommand{\arraystretch}{1.2}
	\setlength{\belowcaptionskip}{4pt}
	\begin{tabular}{|Sc|Sc|Sc|}
		\hline
		Description & Hahne et al. \cite{hahne2016refocusing} & Our\\
		\hline
		\hline
		Main lens focal length &$f_U$& $\fM$\\
		Distance $\Hcam$ to MLA &$b_U$&$\dMm$\\
		Distance $\Hcam$ to virtual image &$b_{U'}$&$\iM$\\
		Distance $\Hscene$ to scene object &$a_{U'}$&$\oM$\\
		Distance MLA to exit pupil &$d_{A'}$& $\dMm - \X$ \\
		Distance MLA to virtual image &$d'_a$& $\dMm - \iM$\\
		ML focal length &$f_s$& $\fm$ \\
		ML diameter &$\Delta s$& $\dML$ \\
		Pixel size &$\Delta u$& $\px$\\
		\hline
	\end{tabular}
	\caption{General notation in Hahne et al. \cite{hahne2016refocusing} and our work.}\label{table:hahne_transfer1}
\end{table}
In addition to the general camera properties listed in \cref{table:hahne_transfer1}, Hahne et al. \cite{hahne2016refocusing} introduce the notation in \cref{table:hahne_transfer2} to exemplarily describe the camera-side rays for a horizontal cross section of the plenoptic camera model.
\begin{table}[h]
	\centering
	\renewcommand{\arraystretch}{1.2}
	\setlength{\belowcaptionskip}{4pt}
	\begin{tabular}{|Sc|Sc|}
		\hline
		Description & Hahne et al. \cite{hahne2016refocusing}\\
		\hline
		\hline
		Horizontal number of microlenses&$J$\\
		Microlens index & $j$\\
		Microlens center of lens $j$ & $s_j=\left(j-\frac{J-1}{2}\right)\cdot\Delta s$ \\
		Microlens image center of lens $j$ & $u_{c,j}=\frac{s_j}{d_{A'}}\cdot f_s+s_j$ \\
		Position of i-th neighbor pixel of MIC & $u_{c+i,j}=u_{c,j}+i\cdot \Delta u$ \\
		\makecell{Slope of ray from i-th\\ neighbor through ML center} &$m_{c+i,j}=\frac{s_j-u_{c+i,j}}{f_s}$\\
		\hline
	\end{tabular}
	\caption{Further notation in Hahne et al. \cite{hahne2016refocusing} without direct equivalents in our setup.}\label{table:hahne_transfer2}
\end{table}
With these the linear ray function originating in a pixel at the sensor position $u_{c+i,j}$ and running through the corresponding microlens center $s_j$ can be described via
\begin{align}
	f_{c+i,j}(z) &= m_{c+i,j} \cdot z + s_j = \frac{s_j-u_{c+i,j}}{f_s} \cdot z + s_j\nonumber\\
	&=\frac{s_j-(u_{c,j}+i\cdot \Delta u)}{f_s} \cdot z + s_j=\frac{s_j-\left(\frac{s_j}{d_{A'}}\cdot f_s+s_j+i\cdot \Delta u\right)}{f_s} \cdot z + s_j.
\end{align}
By assuming an MLA with a microlens center located at the main lens optical axis, the ray originating from the center of the central microlens image going through the microlens center $s_0=0$ can be described by
\begin{align}
	f(z):=f_{c,0}(z) =\frac{s_0-\left(\frac{s_0}{\dMm-\X}\cdot \fm+s_0+0\cdot \px\right)}{\fm} \cdot z + s_0 = 0.
\end{align}
Given a second pixel position at distance $\hMLI+\hat{\shift}$ from the central pixel, this pixel's index in the microlens image $j=1$ is given by $i=\frac{\hat{\shift}}{\px}$. Accordingly, the ray originating from this pixel running through the microlens center $s_1=\dML$ is given by
\begin{align}
	\tilde{f}(z):=f_{c+i,1}(z) &=\frac{s_1-\left(\frac{s_1}{\dMm-\X}\cdot \fm+s_1+\frac{\hat{\shift}}{\px}\cdot \px\right)}{\fm} \cdot z + s_1\nonumber\\ 
	&= \frac{\dML-\left(\frac{\dML}{\dMm-\X}\cdot \fm+\dML+\hat{\shift}\right)}{\fm} \cdot z + \dML\nonumber\\
	&= \frac{\dML-\left(\hMLI+\hat{\shift}\right)}{\fm} \cdot z + \dML.
\end{align}

\subsection[\appendixname~\thesubsection]{Dansereau et al.\cite{dansereau2013decoding}}\label{append:dansnotation}
\begin{table}[h]
	\centering
	\renewcommand{\arraystretch}{1.2}
	\setlength{\belowcaptionskip}{6pt}
	\begin{tabular}{|Sc|Sc|Sc|}
		\hline
		Description & Dansereau et al. \cite{dansereau2013decoding} & Our\\
		\hline
		\hline
		Distance $\Hcam$ to MLA &$d_M$& $\dMm$ \\
		\makecell{Distance sensor to MLA\\(equals ML focal length for SPCs)} &$d_\mu$& $\fm$ \\
		ML focal length &$f_\mu$& $\fm$ \\
		
		\hline
	\end{tabular}
	\caption{General notation in Danserau et al. \cite{dansereau2013decoding} and our work.}\label{table:dans_transfer}
\end{table}
Apart from the intrinsic parameters listed in \cref{table:dans_transfer}, many other aspects of the notation in Danserau et al. \cite{dansereau2013decoding} are dedicated to the light field outside the camera, which is not directly modeled in our work.

\subsection[\appendixname~\thesubsection]{Pertuz et al.\cite{pertuz2018focus}}\label{append:pertuznotation}
\begin{table}[h]
	\centering
	\renewcommand{\arraystretch}{1.1}
	\setlength{\belowcaptionskip}{6pt}
		\begin{tabular}{|Sc|Sc|Sc|}
			\hline
			Description & Pertuz et al. \cite{pertuz2018focus} & Our\\
			\hline
			\hline
			Main lens focal length &$f$& $\fM$\\
			Focus distance (measured from $\Hscene$) & $z_0$ & $\oM_f$ \\
			\makecell{Corresponding image distance\\(equals distance MLA to $\Hcam$)} & $x_0$ & $\dMm$ \\
			Target distance & $z$ & $\oM$\\
			Corresponding image distance & $x$ & $\iM$\\
			Distance MLA to sensor (equals ML focal length) & $\beta$ & $\fm$ \\
			ML diameter & $D$ & $\dML$ \\
			Pixel size & $\mu$ & $\px$ \\
			Separation between real and synthetic focal plane & $\Delta x$ & $F-\Fnew$\\
			Refocusing parameter (sub-aperture image shift) & $\rho$ & $\shiftwrong$ \\
			UV plane step size & $\Delta u$ & $\deltaUV$ \\ 
 			\hline
		\end{tabular}
	\caption{General notation in Pertuz et al. \cite{pertuz2018focus} and our work.}\label{table:pertuz_transfer}
\end{table}
Using the notation as given in \cref{table:pertuz_transfer}, the deduction steps for the model in \cite{pertuz2018focus} can be translated as stated in \cref{table:pertuz_transfer2}.
\begin{table}[h]
	\centering
	\renewcommand{\arraystretch}{1.1}
	\setlength{\belowcaptionskip}{6pt}
	\begin{tabular}{|Sc|Sc|Sc|}
		\hline
		Equation number in \cite{pertuz2018focus} & Pertuz et al. \cite{pertuz2018focus} & Our Notation\\
		\hline
		\hline
		(1) & $\frac{1}{f}=\frac{1}{x_0}+\frac{1}{z_0}$& $\frac{1}{\fM}=\frac{1}{\dMm}+\frac{1}{\oM_f}$\\
		(4) & $\Delta u = \frac{\mu}{\beta D}\Delta x \mathbf{u}$ & $\deltaUV = \frac{\px}{\fm\dML}(F-\Fnew)\cdot u$ \\
		(5) & $\rho = \frac{\mu}{\beta D} \Delta x$ & $-\shiftwrong = \frac{\px}{\fm\dML}(F-\Fnew)$\\
		(7) & $z = \frac{f(x_0-\Delta x)}{x_0-\Delta x - f}$ & $\oM = \frac{\fM(\dMm - (F-\Fnew))}{\dMm - (F-\Fnew)-\fM}$\\
		(8) & $z = z_0\left(\frac{1-a_0\rho}{1-a_1\rho}\right)$ & $\oM = \oM_f\left(\frac{1-a_0(-\shiftwrong)}{1-a_1(-\shiftwrong)}\right)$ \\
		Addition to (8) &$a_0=\frac{\beta D}{\mu x_0}$&$a_0=\frac{\fm\cdot\dML}{\px\cdot\dMm}$\\
		Addition to (8) &$a_1=z_0a_0 / f$& $a_1=\frac{\oM_f\cdot a_0}{\fM}$\\
		\hline
	\end{tabular}
	\caption{Crucial equations from Pertuz et al. \cite{pertuz2018focus} and the corresponding notation in our work.}\label{table:pertuz_transfer2}
\end{table}
Note, however, that this table presents a direct notation transfer, especially regarding equations (5) and (8) of \cite{pertuz2018focus}. As shown in \cref{sec:Revisiting}, the shift parameter $\rho$ is already based on an incorrect assumption and does therefore not directly correspond to $\shiftwrong$.

\section[\appendixname~\thesection]{Thick Microlenses}\label{append:thick}
\begin{figure}[h!]
	\centering
	\includegraphics[page=1]{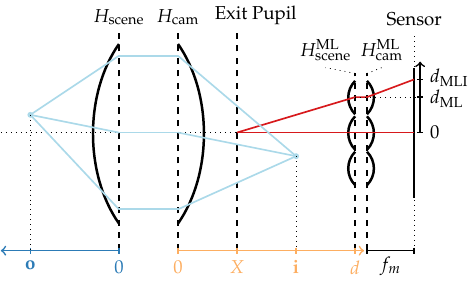}
	\caption{Extension of the plenoptic camera model shown in \cref{fig:lensmodel} by a non-zero microlens thickness.}
	\label{fig:thickmicro}
\end{figure}

In this appendix, the thin microlenses are extended by a model featuring separate principal planes $\HcamML$ and $\HsceneML$ , i.e. a non-zero thickness. The separation between the main lens camera-side principal plane $\Hcam$ and the MLA, $\dMm$, is then given by the distance between $\Hcam$ and $\HsceneML$. On the other side, the distance between MLA and sensor, which equals the focal length of the microlenses, $\fm$, now describes the distance between $\HcamML$ and the sensor plane as shown in \cref{fig:thickmicro}. 
Since a ray entering $\HsceneML$ at a certain height leaves the microlens at the same height at $\HcamML$, the deductions of \cref{sec:SPC_Optics} and \cref{sec:erroranalysis} are still valid.

\section[\appendixname~\thesection]{Evaluation Setups}\label{append:setups}
In the following tables \ref{table:lenses_patents}, \ref{table:lenses_fin} and \ref{table:lenses_inf}, the specific lens models, associated patents and the SPC configurations are listed. 
\begin{table}[h]
	\centering
	
	\renewcommand{\arraystretch}{1.2}
	\setlength{\belowcaptionskip}{6pt}
	
	\resizebox{0.76\textwidth}{!}{%
		\begin{tabular}{|Sc|Sc|Sc|}
			\hline
			Manufacturer & Model & Patent \\
			\hline
			\hline
			Rodenstock & Sironar-N 100mm f5.6 & DE 2729831 Example 1\\
			\hline
			Zeiss & Batis 1.8 / 85& JP 2015-096915 Example 2\\
			\hline
			Ricoh & smc Pentax-A 200mm F4 Macro ED& US 4,666,260 Example 1\\
			\hline
			Canon & EF85mm f1.8 USM& JP 1993-157964 Example 1\\
			\hline
			Olympus& Zuiko Auto-Zoom 85-250mm F5&US 4,025,167 Example 2\\
			\hline
		\end{tabular}
	}
	\caption{Model names of the simulated main lenses and associated patents specifying the lens element parameters. DE, JP and US specify the country of the patent application.\cite{claff}\label{table:lenses_patents}}
\end{table}
\begin{table}[h]
	\centering
	
	\renewcommand{\arraystretch}{1}
	\setlength{\belowcaptionskip}{6pt}
	
	\resizebox{0.76\textwidth}{!}{%
		\begin{tabular}{|Sc|Sc|Sc|Sc|Sc|Sc|}
			\hline
			& \multicolumn{5}{c|}{ Finite Focus}\\
			\cline{2-6}
			Property& Rodenstock & Zeiss & Ricoh & Canon & Olympus\\
			\hline
			\hline
			Focal Length $\fM$ (mm)& 99.998& 82.047&167.994&84.998&85.120\\
			$\Hcam$ (mm)& -1.395 &-30.395&-118.018&-2.303&54.527\\
			$\Hscene$ (mm)&-0.981 &36.192&83.127&-27.317&1.026\\
			Exit Pupil Loc. (mm)&-1.201&10.257&-18.109&-31.241&-5.690\\
			Exit Pupil Radius (mm)& 8.617&2.423&3.822&6.512&2.917\\
			$\X$ (mm)&0.194&40.652&99.909&-28.938&-60.219\\
			$f$-number&5.600&8.500&8.000&7.330&6.590\\
			Focus Distance $F$ (mm)& 500&500&500&300&300\\
			ML Pitch ($\mu m$)&178.158&173.703& 176.246 &177.856&110.567\\
			ML Diam ($\mu m$) & 178.158&173.703& 176.285 &177.856&110.567\\
			ML Focal Length $\fm$ (mm)&1.290&2.084&3.261&1.779&0.884\\
			MLA Thickness (mm) & 0.100&0.100&0.100&0.100&0.100\\
			MLA-Sensor Dist (mm) & 1.280&2.084&3.261&1.779&0.874\\
			Aperture-Sensor Dist (mm) & 124.702&69.944&138.341&118.176&174.140\\
			Sensor Width (mm) & 23.220&23.220&23.220&23.220&7.222\\
			Sensor Height (mm) & 23.220&23.220&23.220&23.220&7.222\\
			\hline
	\end{tabular}}
	\caption{SPC properties for the evaluation setups with finite focus distance.\label{table:lenses_fin}}
\end{table}
\begin{table}[h]
	\centering
	
	\renewcommand{\arraystretch}{1}
	\setlength{\belowcaptionskip}{6pt}
	
	\resizebox{0.76\textwidth}{!}{%
		\begin{tabular}{|Sc|Sc|Sc|Sc|Sc|Sc|}
			\hline
			& \multicolumn{5}{c|}{ Infinite Focus}\\
			\cline{2-6}
			Property& Rodenstock & Zeiss & Ricoh & Canon & Olympus\\
			\hline
			\hline
			Focal Length $\fM$ (mm)& 99.998& 82.860&173.115&84.998&85.004\\
			$\Hcam$ (mm)& -1.395 &-29.315&-109.963&-2.303&54.617\\
			$\Hscene$ (mm)&-0.981 &34.023&75.005&-27.317&1.155\\
			Exit Pupil Loc. (mm)&-1.201&10.257&-18.109&-31.241&-5.690\\
			Exit Pupil Radius (mm)& 8.617&3.817&7.093&6.512&9.290\\
			$\X$ (mm)&0.194&39.572&91.525&-28.938&-60.308\\
			$f$-number&5.600&5.870&9.100&8.000&7.660\\
			Focus Distance $F$ (mm)& $\infty$&$\infty$&$\infty$&$\infty$&$\infty$\\
			ML Pitch ($\mu m$)&178.158&175.944& 177.164 &177.840&171.492\\
			ML Diam ($\mu m$) & 178.158&175.944& 177.164 &177.840&171.492\\
			ML Focal Length $\fm$ (mm)&1.032&0.998&1.230&1.383&1.297\\
			MLA Thickness (mm) & 0.100&0.100&0.100&0.100&0.100\\
			MLA-Sensor Dist (mm) & 1.022&0.987&1.220&1.373&1.287\\
			Aperture-Sensor Dist (mm) & 99.745&54.659&64.521&84.189&141.035\\
			Sensor Width (mm) & 23.220&23.220&23.200&23.220&22.32\\
			Sensor Height (mm) & 23.220&23.220&23.200&23.220&22.32\\
			\hline
	\end{tabular}}
	\caption{SPC properties for the evaluation setups with infinite focus distance.\label{table:lenses_inf}}
\end{table} Note, that $\Hcam$ and the exit pupil location are measured from the main lens aperture position and have a positive sign for these planes being located on the sensor side of the aperture. On the other hand, $\Hscene$ is measured from the aperture with a positive sign if located on the scene-side of the aperture. $\X$ is measured from the position of $\Hcam$ as shown in \cref{fig:lensmodel} and all remaining properties are assumed to be unsigned distance values. All mentioned properties were measured via a 2d ray tracer which we also integrated into the Blender add-on to facilitate fast automatic plenoptic camera reconfigurations.

\begin{adjustwidth}{-\extralength}{0cm}

\reftitle{References}


\bibliography{main}

\PublishersNote{}
\end{adjustwidth}
\end{document}